\documentclass[letterpaper]{article} 
\usepackage{aaai24}  

\usepackage{times}  
\usepackage{helvet}  
\usepackage{courier}  
\usepackage[hyphens]{url}  
\usepackage{graphicx} 
\usepackage{natbib}  
\usepackage{caption} 
\usepackage{algorithm}
\usepackage{algorithmic}

\usepackage{amsmath}
\usepackage{amsfonts}
\usepackage{newfloat}
\usepackage{listings}
\DeclareCaptionStyle{ruled}{labelfont=normalfont,labelsep=colon,strut=off} 
\floatstyle{ruled}
\newfloat{listing}{tb}{lst}{}
\floatname{listing}{Listing}
\title{On the Robustness of Decision-Focused Learning}
\author {
    Yehya Farhat
}
\affiliations{
    Syracuse University\\
    ymfarhat@syr.edu
}

\begin{document}

\maketitle

\begin{abstract}

Decision-Focused Learning (DFL) is an emerging learning paradigm that tackles the task of training a machine learning (ML) model to predict missing parameters of an incomplete optimization problem, where the missing parameters are predicted. DFL trains an ML model in an end-to-end system, by integrating the prediction and optimization tasks, providing better alignment of the training and testing objectives. DFL has shown a lot of promise and holds the capacity to revolutionize decision-making in many real-world applications. However, very little is known about the performance of these models under adversarial attacks. We adopt ten unique DFL methods and benchmark their performance under two distinctly focused attacks adapted towards the Predict-then-Optimize problem setting. Our study proposes the hypothesis that the robustness of a model is highly correlated with its ability to find predictions that lead to optimal decisions without deviating from the ground-truth label. Furthermore, we provide insight into how to target the models that violate this condition and show how these models respond differently depending on the achieved optimality at the end of their training cycles.

\end{abstract}

\section{Introduction}

\textit{Decision Focused Learning} (DFL) is an emerging learning paradigm in the field of \textit{Machine Learning} (ML) that confronts the task of decision-making under uncertainty. The problem of decision-making under uncertainty pertains itself to the sequential task of predicting uncertain quantities using an ML model and then using a \textit{Constrained Optimization} (CO) model to optimize objectives using the predicted quantities.

The conventionally adopted approach to this problem is to consider both the predictive and optimization tasks separately. This \textit{Two-Stage} (TS) approach first involves training an ML model using a traditional loss function such as MSE to map the input features to the relevant parameters of the CO problem. Where afterwards a specialized optimization algorithm is used to solve the CO problem. This approach has obvious advantages, in which the model learning phase is well justified and independent of any secondary task besides finding the best mapping between the input features and labels. In theory, this approach is infallible if we are able to obtain a perfect model to make precise predictions. Indeed if the predictions of the parameters by the trained model are perfectly accurate, they would lead to correct specifications of the CO models which can be solved to yield fully optimal solutions. However, in practice ML models fall short of perfect accuracy which lead these errors propagating to the CO models and which in turn lead to sub-optimal decisions. This put forward the need of modelling the predictive and decision process jointly.  
DFL is based on training the ML model to make predictions that lead to good decisions. DFl integrates prediction and optimization in an end-to-end system that targets predictions by optimizing a loss function based on the subsequent decisions. DFL has been shown to outperform TS on a variety of domains \cite{hardSOP, wilder2018melding, cameron2021perils}.

The key technical challenge with the DFL paradigm is the gradient-based training of the ML model. Given that the loss needs to be differentiated through the entire prediction and optimization pipeline, a discontinuous mapping is produced when the CO problem operates on discrete variables. To overcome these challenges, many smooth surrogate models for these discrete mappings, along with their differentiation, have been proposed. For an extensive survey of the different methodologies, interested readers are referred to \cite{mandi2023decisionfocused} 

DFL has garnered increasing attention, but the robustness of these pipelines in adversarial settings remains inadequately understood. Robustness in a model refers to its ability to maintain performance when subjected to adversarial perturbations or novel data inputs. To employ these pipelines effectively in real-world contexts, comprehending their robustness is crucial, as it allows us to evaluate the suitability of the models in various situations.
We analyze a model's robustness by introducing small worst-case perturbations to the input and evaluate their impact on the model’s overall downstream performance. The code and data used are accessible through \textbf{https://github.com/yehya-farhat/AdvAttacks-DFL.git}. To the extent of our knowledge, this work is the first to analyze and compare the robustness of the different DFL methodologies under such an adversarial settings.\\
Overall, this paper makes the following  contributions: 
\begin{itemize}
  \item We replicate eleven DFL methods and implement two types of adversarial attacks with proper adaptation to cater towards Predict-then-optimize based problems 
  \item We comprehensively evaluate the DFL methods across three distinct problems and summarize the results  
  \item We offer empirical insights to demonstrate that robustness is highly correlated with the models ability to find the optimal solutions with respect to the ground truth labels
\end{itemize}

\section{Related Work}
Model robustness is a crucial concept extensively examined within the realms of deep learning and adversarial machine learning. A multitude of studies have delved deeply into understanding and enhancing model robustness, highlighting the vulnerability of neural networks to specially crafted inputs.
\textit{Evasion attacks} represent a methodology where specific inputs are meticulously designed to mislead a neural network, thereby undermining or altering the model’s output. Such vulnerabilities have been meticulously documented and analyzed in previous works \cite{goodfellow2015explaining, kurakin2017adversarial}. Another subtle yet potent form of undermining model robustness is through \textit{poisoning attacks}. In this approach, the adversary introduces manipulated data points into the training dataset, intending to degrade the model’s overall performance and reliability.\\
Despite the success and extensive study of evasion attacks, their implementation and performance in non-classification problem settings remains understudied. Regression tasks have no natural margins as in the case of classification tasks that lend themselves to the nice property of having finite output spaces. Adversarial attacks in regression settings are hindered by challenges in defining success margins and establishing evaluation metrics. Despite these challenges, strides have been made in generalizing evasion attacks for regression problems. \cite{tong2018adversarial} looked at attacks in a special case of ensemble linear models, although insightful this linear case doesn't capture the larger class of non-linear regression problems. Other studies have focused on specific applications of regression problems. \cite{ghafouri2018adversarialCPS}
Study the regression task in cyber-physical systems by selecting an optimal threshold for each sensor against an adversary. \cite{deng2020analysisAD} Examine the regression task of predicting the optimal steering angle in autonomous driving by using a concept they called an adversarial threshold. They consider an attack successful if the deviation between the original prediction and the prediction of an adversarial example is below the specified threshold. \cite{meng2019whiteboxBCI} Introduce two attacks for specific regression problems in the field of electroencephalogram based brain-computer interfaces and utilize a similar attack success criteria as above. \cite{gupta:hal-03527640} Propose a flexible adversarial attack for regression tasks and analyze its effectiveness using various error metrics.  
Furthermore, general defenses have been proposed in the context of regression problems. \cite{nguyen2018adversarial} introduce a useful defense to reduce the effectiveness of evasion attacks by considering adversarial attacks as a symptom of numerical instability of the learned weight matrix.\\
In the context of DFL, the exploration of adversarial attacks is an emerging area of study. It has been demonstrated that effective poisoning attacks can be generated against DFL models \cite{DFLpoisoning}. 
Concurrently, a body of work has emerged that focuses on bolstering DFL robustness against label noise. \cite{Yu2023Tambe} Cast the learning problem as Stackelberg games and provide bounds on decision quality in the presence of adversarial label drift and propose a learning schema to mitigate this affect by anticipating label drift.

\section{Background}
\subsection{Predict-Then-Optimize Problem Setting}
Decisions are often mathematically modeled using CO problems. These problems are well-suited to a wide range of decision-making scenarios. In many real-world applications, certain parameters of the CO problem remain uncertain and require inference from contextual data. The Predict-and-Optimize problem setting involves decision models that are represented by partially defined optimization problems, whose specification is completed by parameters that are predicted from data.
\begin{subequations}
\begin{alignat}{2}
& x^*(\mathbf{c}) &&= \arg\min_x\ f(x, \mathbf{c}) \label{eq:1a} \\
     &  \text{s.t.} \quad && g(x) \leq 0 \label{eq:1b} \\
    & && h(x) = 0. \label{eq:1c}
\end{alignat}
\end{subequations}
The goal of the optimization problem described is to find a minimizer $x^{*}(\mathbf{c}) \in \mathbb{R}^p$ of the objective function (1a), satisfying the constraints (1b-1c). 
This paper considers the problem setting where the parameter $\mathbf{c} \in \mathcal{C}  \subseteq \mathbb{R}^k $ is unknown and must be inferred as a function of observed empirical features $\textbf{z} \in \mathcal{Z} \subseteq \mathbb{R}^l$.
In this context, given a set of past observations pairs $\mathcal{D}_{train} = \{ (\mathbf{z_i}, \mathbf{c_i})  \}^{N}_{i=1}$ The predictive task is to learn a parameterized (by $w$) model $m_{w} : \mathcal{Z} \rightarrow  \mathcal{C} $. Such that the model is used to make predictions in the form of $\hat{\mathbf{c}} = m_{w}(z)$. Subsequently, a decision $x^{*}(\mathbf{\hat{c}})$ is made based on the predicted parameter. The overarching learning objective is to optimize the set of decisions made over the set of observed features  $\mathbf{z} \in \mathcal{Z}$, with respect to some evaluation criterion on those decisions.   

\subsection{Frameworks for Predict-Then-Optimize}
While the ML model $m_{w}$ is trained to predict $\mathbf{\hat{c}}$, its performance is evaluated on the basis of the corresponding optimal solution of the decision model $x^*(\mathbf{\hat{c}})$. Using standard ML approaches, learning the model $m_{w}$ can only be supervised by ground-truth $\mathbf{c}$ using a standard loss function $\mathcal{L}(\mathbf{\hat{c}},\mathbf{c})$, such as mean squared error (MSE). But given that we are concerned with optimizing the decision $x^{*}(\mathbf{\hat{c}})$, it would be favorable to train $m_{w}$ to make predictions $\mathbf{\hat{c}}$ that optimize the the decision. 
This puts forth two different learning approaches for the model $m_{w}$. 

\subsubsection{Two-Stage (TS)}
The conventional learning paradigm is to generate accurate parameter predictions $\mathbf{\hat{c}}$ with respect to the ground-truth labels $\mathbf{c}$, this Two-Stage approach learns the predictive model by minimizing a standard loss function, such as MSE.
\begin{equation}
\label{eq:twostage}
\min_{w} \sum_{(z,c) \in \mathcal{D}_{\text{train}}} \|m_{w}(z) - c\|^2
\end{equation}
The final trained model is then used to make a prediction on a new data point $\mathbf{z}$ such that the prediction is used to optimize the objective $x^{*}(m_{w}(\mathbf{z}))$ in Eq. (\ref{eq:1a})

\subsubsection{Decision-Focused learning (DFL)}
By contrast, DFL trains the ML model to optimize the evaluation criteria that measures the quality of the decision. 
\begin{equation}
\label{eq:DFL}
\min_{w} \sum_{(z,c) \in \mathcal{D}_{\text{train}}} f(x^{*}(m_{w}(z)),c) - f(x^{*}(c),c)
\end{equation}
Similar to the TS approach, after training the model is called  to make new predictions on new data points, which are then passed to (\ref{eq:1a}). The advantage of DFL is the alignment of the training objective and the testing objective $f$. To optimize the objective in DFL, it is common to use gradient descent. This will require backpropagating through the optimal decision $x^*$. Depending on the CO problem, The gradient of the optimal decision is not always easily computed. CO problems can be categorized in terms of the form taken by the their objectives and constraints. These forms define the properties of the optimization mapping, such as its continuity and differentiability. This presents two major challenges: firstly, the optimization problem's solution mapping $\mathbf{\hat{c}} \rightarrow \mathbf{x^{*}(\hat{c})}$ lacks a closed form that can be differentiated directly. Secondly, depending on the problem, many useful optimization problems exhibit non-differentiable mappings at some points and zero-valued gradients at others, rendering gradient descent unusable. These challenges highlight the need for developing methodologies that address the difficulty of differentiating an optimization mapping in DFL during gradient-based training. Various approaches suggest different smoothed surrogate approximations suitable for backpropagation.

\section{Methodology}
\subsection{Objective} 
The problem of adversarial attacks is closely related to the robustness issue for a neural network, i.e. its sensitivity to perturbations. Let $m : \mathcal{Z}  \rightarrow \mathcal{C}\ $ denote the trained model, for which either Eq. (\ref{eq:twostage}) or Eq. (\ref{eq:DFL}) was used for training to produce an output $m(z)$. We define an additive vector $\epsilon \in \mathbb{R}^l$ to perturb the original input and get the perturbed output $m(z + \epsilon)$. Unlike classification problems the output of the model does not have well defined success margins, in the classification scenario we desire no change in the response for small change in the input. For regression problems, we must expect \textit{some} change in the response for any change in the input. In such a setting, attacking the network amounts to finding a perturbation $\epsilon$ of preset magnitude which makes the output maximally deviate from a reference output. The reference output may be defined as the model output $m(z)$ or the ground truth label $c$. In this work we are interested in studying the behavior of the different DFL methodologies under input perturbations, thus we will also adopt the reference output of the decision problem under the model prediction $x^{*}(m(z))$ and the ground truth label $x^{*}(c)$. In this context, the measure of deviation and magnitude of perturbation play an important role in the problem formulation. We measure the output deviation using an  $\ell_{q}$-norm where $q \in [1,+ \infty]$. It is important to underscore that opting for this approach is logical in the case of regression problems. The $\ell_2$ and $\ell_1$ norms are often applied as loss functions during the training process. Conversely, the $\ell_{+\infty}$ norm is commonly utilized when addressing matters of reliability  

Given a robust predictive function ${m_{r}(.)}$ and a quality function $f(.)$, we would expect that 
\begin{equation}
\label{eq:offsetQuality}
 |f(m_{r}(z)) -  f(m_{r}(z + \epsilon))| \leq \Delta  
\end{equation}
\begin{equation}
 \label{eq:offsetPred}
 |m_{r}(z) -  m_{r}(z + \epsilon)| \leq \delta  
\end{equation}
for some threshold $\Delta$ and $\delta$. Most adversarial work focuses on attacks that violate this expectation. Making it such that small perturbations cause drastic changes in the output. Its important to note that satisfying both Eq. (\ref{eq:offsetQuality}) and Eq. (\ref{eq:offsetPred}) need not be necessary in the context of DFL. DFL focuses on making predictions that lead to good quality decision, regardless of the prediction itself. In cases where the CO problem has non-unique optimal solutions, DFL might allow for the satisfaction of Eq. (\ref{eq:offsetQuality}), while simultaneously allowing for the violation of Eq. (\ref{eq:offsetPred}). But in a setting where we have a perfect predictive and robust model (with respect to the ground truth labels) then we would expect both inequalities to be satisfied. 

In this work we focus on studying the behavior of $\Delta$ and $\delta$ with respect to the different DFL methodologies. Our objective is to categorize the sensitivity of each learned model by measuring the variations in output under two distinctly focused attacks. 
Additionally, we analyze the relationship between the behavior of the downstream task $f(m_{r}(z + \epsilon))$ and the predictive task $m_{r}(z + \epsilon)$, aiming to understand how perturbations affect both tasks.

\subsection{Problem Sets}
\subsubsection{Problem description} 
We select 3 diverse problems for benchmarking: Warcraft shortest path, Portfolio optimization, and Knapsack. All of which have been previously used as benchmarks in the DFL literature and their datasets are publically available. All these problems pose both a predictive and optimization task.


\subsubsection{Warcraft Shortest Path}
This problem was adopted from the work of \cite{DBLP}, utilizing the openly accessible warcraft terrain map images dataset \cite{war2edit_2023}. It features images configured as a $d \times d$ grid, with each grid cell—or pixel—assigned a specific cost that requires prediction. The objective is to then identify the shortest path from the top left-pixel to the bottom-right pixel. Unless situated on the grid's boundary, one may proceed in any of eight possible directions from a given pixel, framing the challenge as a node-weighted shortest path problem on a graph with $d^2$ vertices and up to $d^2$ edges. This problem is transformed into the more conventional edge-weighted shortest path problem by dividing each weighted node into a pair of nodes, with the original node's weight transferred to the edge that connects these new nodes. \\
The problem of shortest path can be formulated as an LP problem with the following form: 

\begin{subequations}
\begin{alignat}{2}
 & \min_x\ c^{\top}x \label{eq:2a} \\
     & \:  \text{s.t.} \quad  Ax = b \label{eq:2b} \\
     & \quad \quad \: \: 0 \leq x \label{eq:2c}
\end{alignat}
\end{subequations}
Where $A \in \mathbb{R}^{|V| \times |E|}$ is the incidence matrix of the graph. $x \in \mathbb{R}^{|E|}$ is a binary vector whose entries are 1 if the corresponding edge is selected and 0 otherwise. $b \in \mathbb{R}^{|V|}$ is a vector whose entries are all 0 except the entries that corresponding to the source and sink node where are 1 and -1, respectively.

The predictive task involves the use of a \textit{convolutional neural network} (CNN) to determine the cost associated with each node (pixel), with costs ranging from 0.8 to 9.2 based on the pixel's visible features. The model processes the  $d \times d$  image to output the costs for all $d^2$
pixels. 
\subsubsection{Portfolio Optimization}
This problem was adopted from the work of \cite{elmachtoub2020smart}. The problem entails predicting asset prices based on empirical data, subsequently a risk constrained optimization problem is solved to obtain a portfolio that maximizes expected return. 
The problem formulation of Portfolio optimization can expressed as the following:  
\begin{subequations}
\begin{alignat}{2}
 & \max_x\ c^{\top}x \label{eq:3a} \\
     & \: \text{s.t.} \quad  x^{\top} \Sigma x \leq \lambda \label{eq:3b} \\
     & \quad \quad \: \: \mathbf{1}^{\top}x \leq \mathbf{1} \label{eq:3c} \\ 
     & \quad \quad \: \: 0  \leq x \label{eq:3d}
\end{alignat}
\end{subequations}
The Synthetic input-target pairs $(\mathbf{z},\mathbf{c})$ are randomly generated according to a random function with a specified degree of nonlinearity, we refer to this as $Deg \in \mathbb{N}$. A detailed breakdown of the random function used for input-target generation can be found in the appendix.\\
$\mathbf{1}$  represents a vector composed entirely of ones. $\Sigma$ refers to a pre-established covariance matrix between asset returns.
The predictive task utilizes a simple linear neural network model, whose inputs is a feature vector $\textbf{z} \in \mathbb{R}^l$ and output is the return vector $\mathbf{c}  \in \mathbb{R}^k$ that represents asset prices.

\subsubsection{Knapsack}
This experiment setup was adopted from the work of \cite{hardSOP}. The problem involves solving the knapsack problem's objective, which entails selecting a subset of items with maximum value from a specified set, while adhering to the capacity limitation. The problem formulation can be written as follows: 
\begin{subequations}
\begin{alignat}{2}
 & \max_x\ c^{\top}x \label{eq:4a} \\
     & \: \text{s.t.} \quad  w^{\top} x \leq Capacity \label{eq:4b} \\
     & \quad \quad \: \: x \in \{0,1\} \label{eq:4c}
\end{alignat}
\end{subequations}
The weights of all items and the capacity constraint are known, hence The prediction task is to predict the value of each item. 
The problem dataset utilizes the \textit{Irish Single Electricity Market Operator} (SEMO) \cite{KnapsackDataset}.
In this arrangement, each day is treated as an individual optimization problem, with every half-hour representing an item in the knapsack. Consequently, both the cost vector 
$\mathbf{c}$ and the weight vector $\mathbf{w}$ consist of 48 elements, each corresponding to a half-hour segment. Each item in the cost array is associated with an 8-dimensional feature vector. This vector represents various attributes, encompassing elements like weather conditions and projected energy load.
The weight array remains constant throughout and is synthetically generated, as previously established by \cite{hardSOP}. A detailed breakdown of the synthetic weight vector generation can be found in the appendix.
As the previous problem, the predictive task entails a linear model, whose input is a feature vector $\textbf{z} \in \mathbb{R}^{48 \times 8}$ and output vector $\mathbf{c}  \in \mathbb{R}^{48}$.

\subsection{Attacks}
\subsubsection{Fast Gradient Sign Method: Prediction-Focused}
The Fast Gradient Sign Method (FGSM) is a $L_{\infty}$ bounded attack \cite{goodfellow2015explaining}. Given a training data point $(z,c)$, a cost function $J$, and the model parameters $w$, the adversarial data point $ \Tilde{z} $ is computed as follows
\[\Tilde{z} = z + \epsilon sign(\nabla_{z}J(w,z,c) )\]
where the cost function $J$ is a standard loss that calculates the error of the prediction with respect to the ground truth labels $c$, and where $\epsilon$ is the attack magnitude.

\subsubsection{Fast Gradient Sign Method: Decision-Focused}
Similar to the Prediction-Focused FGSM attack implementation, We implement a slightly modified version of the attack by opting for a different cost function. Let $\acute{J}$ represent the decision cost function that calculates the decision error of the prediction with respect to the ground truth decision $x^{*}(c)$ (Regret). Similarly, we calculate the adversarial data point $ \Tilde{z} $ as   
\[\Tilde{z} = z + \epsilon sign(\nabla_{z} \acute{J} (w,z,c) )\]
Its important to note that the calculation of $\nabla_{z} \acute{J} (.)$ is not always straightforward, as it will run into the same issues of non-differentiability, and zero-valued gradients. To that end, we will need to instead utilize the smoothed surrogate approximations of each respective end-to-end DFL pipeline to compute this attack. 

\subsubsection{Attack Setting}
In our experiments, we assess the performance of the prediction-focused FGSM (PF) and decision-focused FGSM (DF) attacks by varying the attack magnitude parameter $\epsilon$, at levels of $0.01, 0.1,$ and $0.15$, respectively. For each respective DFL methodology, we adopt the same approximation used in training to allow for useful gradient computations of $\nabla_{z} \acute{J} (.)$.    

\begin{table}[t]
\centering

\caption{Overview of the different DFL methodologies adopted for experimentation}
\label{methods}
\begin{tabular}{p{4.5cm} p{3cm}} 
 \hline
 Methodology & Problem Form  \\ 
 \hline\hline
 Quadratic Programming Task Loss (QPTL) \cite{wilder2018melding} & LPs, ILPs \\ 
 Differentiation of Blackbox Combinatorial Solvers (DBB) \cite{DBLP} & Linear Objective \\ 
 Fenchel-Young loss (FY) \cite{blondel2020learning} & Linear Objective \\
 Implicit maximum likelihood estimation (IMLE) \cite{IMLE} & Linear Objective \\
 Interior Point Solving Method (IntOpt) \cite{Intopt} & LPs, ILPs \\
 Smart Predict Then Optimize (SPO) \cite{elmachtoub2020smart} & Linear Objective \\
 Maximum A Posteriori approximation (MAP) \cite{mulamba2021contrastive} & General optimization problems \\ 
 Pairwise Learning to Rank (Pairwise) \cite{mandi2022DFRanking} & General optimization problems \\
 Pairwise Learning to rank Difference (Pairwise Diff) \cite{mandi2022DFRanking} & General optimization problems \\
 Listwise Learning to rank (Listwise) \cite{mandi2022DFRanking} & General optimization problems \\
 \hline
\end{tabular}
\end{table}

\subsection{DFL methodologies}
We adopt $10$ different DFL methodologies and the two-stage method as our baseline for comparison. All mentioned DFL techniques are an attempt to overcome the challenge of differentiating the optimization mapping by proposing different smoothed surrogate approximations of $\frac{d\mathcal{L}(x^{*}(\hat{c}))}{d\hat{c}}$ or $\frac{dx^{*}(\hat{c})}{d\hat{c}}$. Table \ref{methods} provides an overview of the different methods and their form adopted for comparison.

\subsection{Network Architecture and Training}
The network architecture for the Warcraft shortest path is a CNN. As in \cite{mandi2023decisionfocused} we adapted the ResNet18 architecture, which includes the first five layers of ResNet18 followed by a max-pooling operation that assists in predicting the underlying cost for each pixel, and a Relu activation function to ensure positive edge weights. 
In addressing the Knapsack and Portfolio problems, we employ a single-layer feed-forward neural network, devoid of hidden layers (A linear model). The rational behind utilizing such a simple predictive model lies in evaluating the effectiveness of the DFL methods under conditions where predictions lack high accuracy. We utilize Adam as our optimization algorithm \cite{kingma2017adam} and ReduceLROnPlateau learning rate scheduler \cite{pytorch}. We utilize Cvxpylayers \cite{cvxpylayers} as our solver for QPTL, QP problems. For the other methods we employ Gurobi \cite{gurobi} and OR-Tools \cite{ortools} as our solvers. The best hyperparameter selection of each model for each experiment can be found in the appendix.

\begin{table*}[t]
    \centering
    \begin{align*}
    \hline \hline \\
        \text{Mean Accuracy Error (MAE)}        & = \hspace{7em} \: \frac{1}{\mathcal{D}_{\text{test}}} \sum_{i=1}^{\mathcal{D}_{\text{test}}} \|m(z_i + \epsilon) - c_i \|_q \\
        \\
        \hline \\
        \text{Fooling Error (FE)}               & = \hspace{7em} \: \frac{1}{\mathcal{D}_{\text{test}}} \sum_{i=1}^{\mathcal{D}_{\text{test}}} \lvert m(z_i + \epsilon) - m(z_i) \rvert \\
        \\
        \hline \\
        \text{Relative Regret Error (RRE)}      & = \hspace{6em} \: \frac{1}{\mathcal{D}_{\text{test}}} \sum_{i=1}^{\mathcal{D}_{\text{test}}} \frac{c_{i}^{\top}(x^{*}m(z_i + \epsilon) - x^{*}(c_i)) }{c_{i}^{\top}x^{*}(c_i)} \\
        \\
        \hline \\
        \text{Fooling Relative Regret Error (FRRE)} & = \: \frac{1}{\mathcal{D}_{\text{test}}} \sum_{i=1}^{\mathcal{D}_{\text{test}}}  \lvert  \frac{c_{i}^{\top}(x^{*}m(z_i + \epsilon) - x^{*}(c_i)) }{c_{i}^{\top}x^{*}(c_i)} - \frac{c_{i}^{\top}(x^{*}m(z_i) - x^{*}(c_i)) }{c_{i}^{\top}x^{*}(c_i)} \rvert \\
        \\
    \hline \hline
    \end{align*}
     \caption{Error Metrics Considered for Evaluation}
     \label{errorMetrics}
\end{table*}

\subsection{Evaluation Metrics}
To understand and analyze the performance of the DFL methodologies under the proposed adversarial attacks, we train the models using the best hyperparameters combinations for each test problem, 10 trails of all the methods are run on the test dataset. Unless otherwise mentioned the evaluation is done based on the 4 error metrics in Table \ref{errorMetrics}. In practical scenarios, instances in which the CO problem presents multiple non-unique solutions can lead a machine learning model to incorrectly predict all cost parameters, $c$ or $\hat{c}$, as zero during early training. This results in technically optimal but meaningless solutions, thereby reducing interpretability and usefulness. We assume that the optimal solution, $x^{*}(c)$, is derived from an optimization oracle equipped with a predefined method to break ties in cases of non-unique solutions.

\section{Results and Discussion}
In the next section, we assess the average performance of 11 methods in 10 trials against various adversarial attacks. The primary objective of our research is to examine the overall robustness of these models. Consequently, when we mention the 'best' models, we refer to those demonstrating the least deviation from the initial solution when subjected to both types of attacks. Note that for the Two-Stage method, we only present metrics related to the Prediction-Focused attack given that the model does not have a direct surrogate decision loss function associated with it.

\subsection{Knapsack Results}
\begin{figure}[t]
\centering
\includegraphics[width = 8cm, height = 5cm]{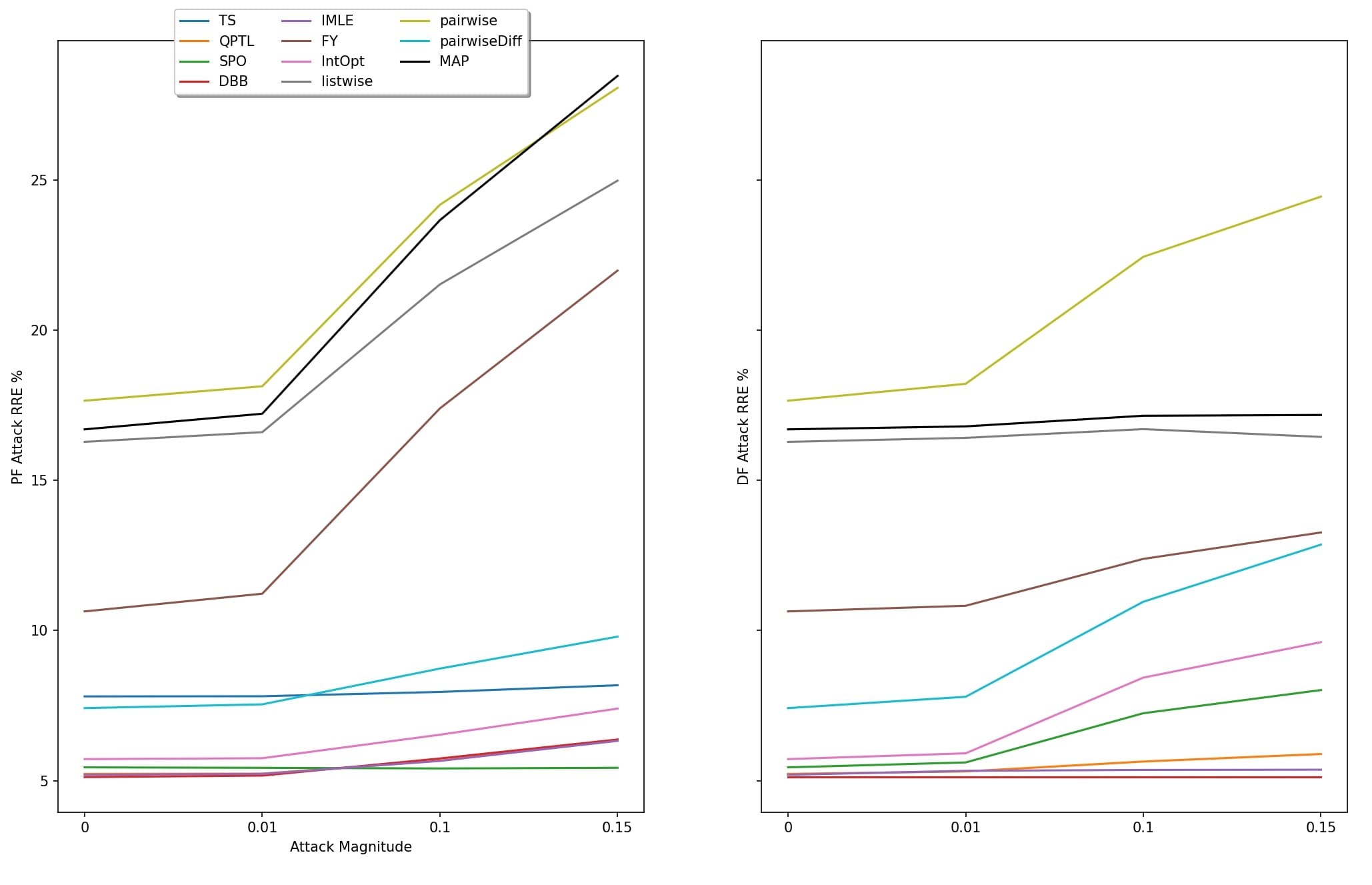} 
\caption{Knapsack capacity 120 RRE, PF Attack left, DF attack right}
\label{Knap120_RRE}
\end{figure}

Two instances of the knapsack problem are under consideration, each associated with a different capacity: 60 and 120. The line plot of relative regret error with respect to the varying attack magnitudes of the problem instance with a capacity of 120 is presented in figure \ref{Knap120_RRE}. 
The generated box plots for the four metrics in Table. \ref{errorMetrics} and the rest of the line plots can be found in the appendix. 

With a capacity of 120, the models exhibiting the best overall robustness on decision quality, are QPTL, TS, DBB, and IMLE. SPO does relatively well under the PF attack but is not able to hold the same performance under the DF attack. Under prediction quality metrics, the best overall performing models are FY, listwise, pairwise, and MAP. With a capacity of 60, the best-performing models in terms of decision quality are DBB and IMLE. FY performs well under the DF attack but poorly under the PF attack. SPO faces a similar issue as observed in the 120-capacity instance: it performs poorly under the DF attack but well under the PF attack. Regarding prediction metrics, the best models are FY, Listwise, Pairwise, and MAP.

There are interesting observations to note on this problem. The first is that the problem clearly has non-unique optimal solutions, as evidenced by all the models exhibiting high MAE errors while tending toward an optimal decision error. The second observation concerns the varied behavior of these models in terms of robustness when subjected to different types of attacks. The models with the highest initial RRE are the least robust under the PF attack while the models with the lowest initial RRE are the most robust under the PF attack. This behaviour is reversed when the models are exposed to the DF attack. Under the DF attack the models exhibiting the lowest initial RRE are the least robust, while the models with highest initial RRE become the most robust.
We believe this is due to what the learned models accomplish at the end of their training cycles and how the different attacks uniquely interact with each method.
By definition the models with the highest RRE have not learned how to make good predictions with respect to the decision. Thus, When these models are faced with attacks that try and maximize prediction error the attack results in the model outputting predictions with no consideration for the decision quality. On the other hand the models with low initial RRE are able to maintain their decision quality due to them having learned how to make good predictions with respect to the decision, irrespective of worst case perturbation with the aim of maximizing prediction error. We hypothesize that the reversal of this behaviour under the DF attack is due to the methods ability to find the optimal solution. When faced with the DF attack most of the models exhibiting the best robustness on the PF attack become the least robust. Assuming that the optimal solution has been found by those methods, maximizing the decision error becomes more feasible. While on the other hand, the models that have not been able to find the optimal solution exhibit relatively small change in decision quality under the DF attack, due the fact that these models have a worse sense of where the optimal solution is.

\subsection{Portfolio Results}
We examine two versions of the Portfolio optimization problem: Degrees 1 and 16, with a noise magnitude parameter set at 1. It's important to note that the IntOpt method is unsuitable in this case because the problem involves quadratic constraints, which the IntOpt method cannot handle. In this problem, for some instances, all the return values are negative, making portfolio optimization with zero return optimal. In such cases, the RRE metric is undefined, as the denominator is zero. Hence, for this problem set, we instead report the absolute regret: $c^{\top}(x^{*}(m(z)) - x^{*}(c))$. We present the line plot of the degree 16 problem instance for the Absolute RE with respect to the attack magnitudes in figure \ref{Port16_AbsRE}. The rest of the line plots and all the box plots can be found in the appendix.
 For this problem, all models perform relatively well, with very negligible variation in decision quality. In such a case the Absolute FRE provides a better glimpse into the minor changes under different attack magnitudes.

\begin{figure}[t]
\centering
\includegraphics[width = 8cm, height = 5cm]{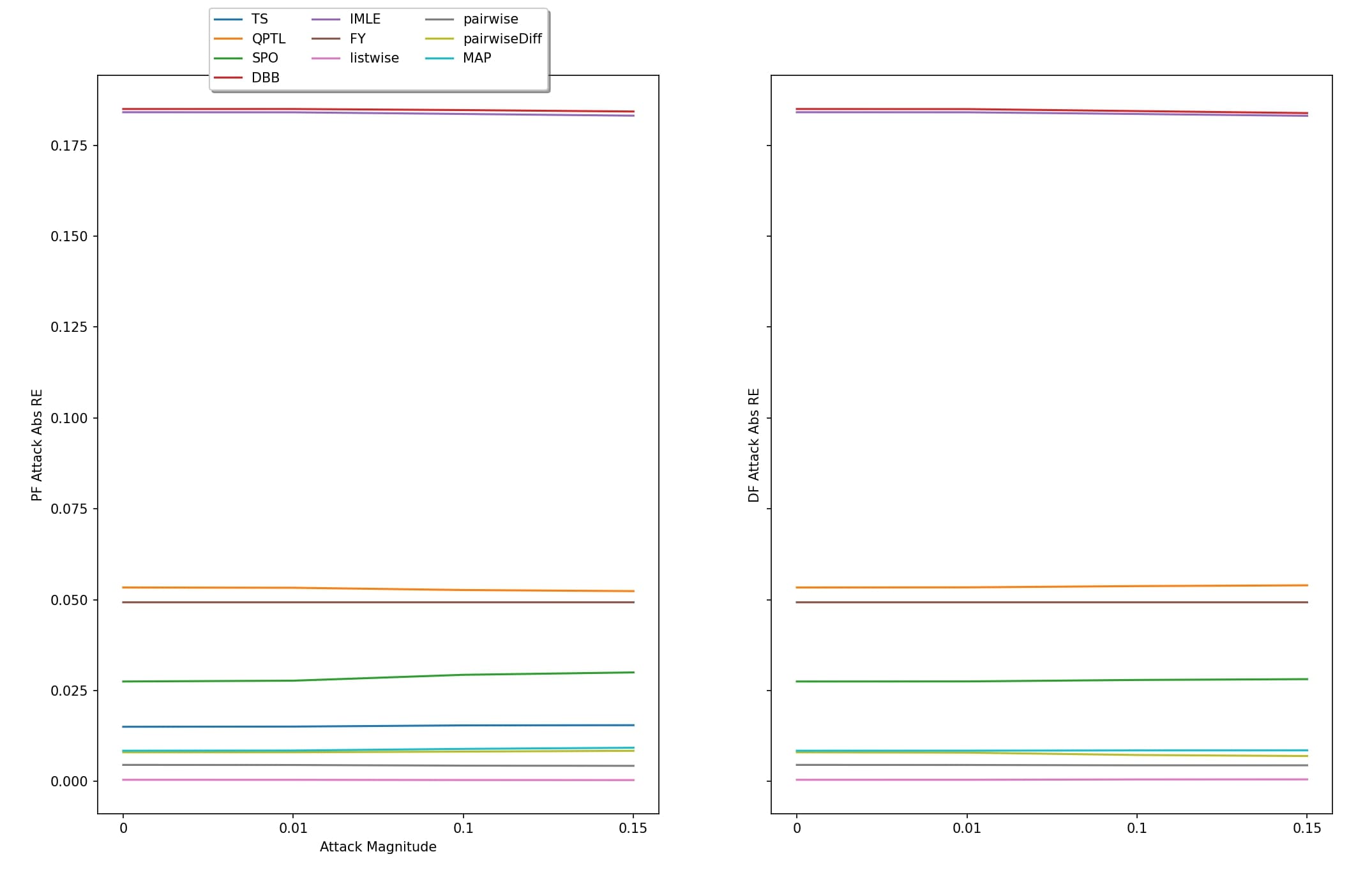} 
\caption{Portfolio deg 16 Absolute RE, PF Attack left, DF attack right}
\label{Port16_AbsRE}
\end{figure}

For the Degree 16 problem instance, the models that exhibit the least amount of variation under all perturbation levels in terms of decision quality are FY and Listwise. With respect to prediction quality, the best-performing models are FY, Listwise, and MAP.
For the Degree 1 problem instance, the best-performing models across all perturbation levels in terms of decision quality are DBB, IMLE, and FY. With respect to prediction quality, the best models are FY and MAP, with FY exhibiting a notably large interquartile range. 

In addressing this problem, we observe an interesting phenomenon: the models with low RE do not demonstrate the expected (As in the knapsack problem) poor robustness when subjected to the DF attack. All models maintain high levels of robustness against both types of attacks. This resilience, we believe, stems from the models successfully identifying the optimal solution based on the ground truth label $c$. This corroborated by the fact that all models consistently exhibit very low MAE, effectively preserving their performance under both attack scenarios.

\subsection{Warcraft Shortest Path}
The evaluation of the warcraft shortest path problem for image sizes of $12 \times 12$ and $24 \times 24$ are presented in the following section. We present the line plot of the RRE with respect to the attack magnitudes for the $24 \times 24$ problem instance in figure \ref{War24_RRE}. The box plots and the rest of the line plots can be found in the appendix. We note, IntOpt and QPTL necessitate the problem's formulation as a Linear Program (LP) and the use of a primal-dual solver. In our experiment, the predictive machine learning model, specifically a CNN, is tasked with predicting the cost associated with each pixel. Training this ML model proves to be a formidable task due to its extensive parameter set. Consequently, integrating this model with computationally demanding components like an interior point optimizer introduces substantial challenges. Owing to these computational constraints, we were unable to conduct experiments using IntOpt and QPTL.

On the Warcraft problem all models exhibit relatively good robustness. 
On the $24 \times 24$ Image size instance, the best overall performing models under decision quality are SPO, IMLE and DBB. Under prediction quality the best models are FY and MAP. On the $12 \times 12$ problem instance, the best models are the same as the $24 \times 24$ problem instance. Under decision quality SPO, DBB, and IMLE are overall the best performers while under prediction quality the best overall models are FY and MAP. 

We notice a parallel with the portfolio problem in our observations: all models demonstrate commendable robustness against both types of attacks while maintaining a low MAE. This lends credence to our hypothesis that a model's decision robustness is directly linked to its ability to identify the optimal solution in relation to the ground truth label $c$. This is particularly evident in the performance of the MAP method. It emerges as the model most adversely affected under both adversarial attacks and simultaneously records the highest MAE, thereby reinforcing our hypothesis. 

\begin{figure}[t]
\centering
\includegraphics[width = 8cm, height = 5cm]{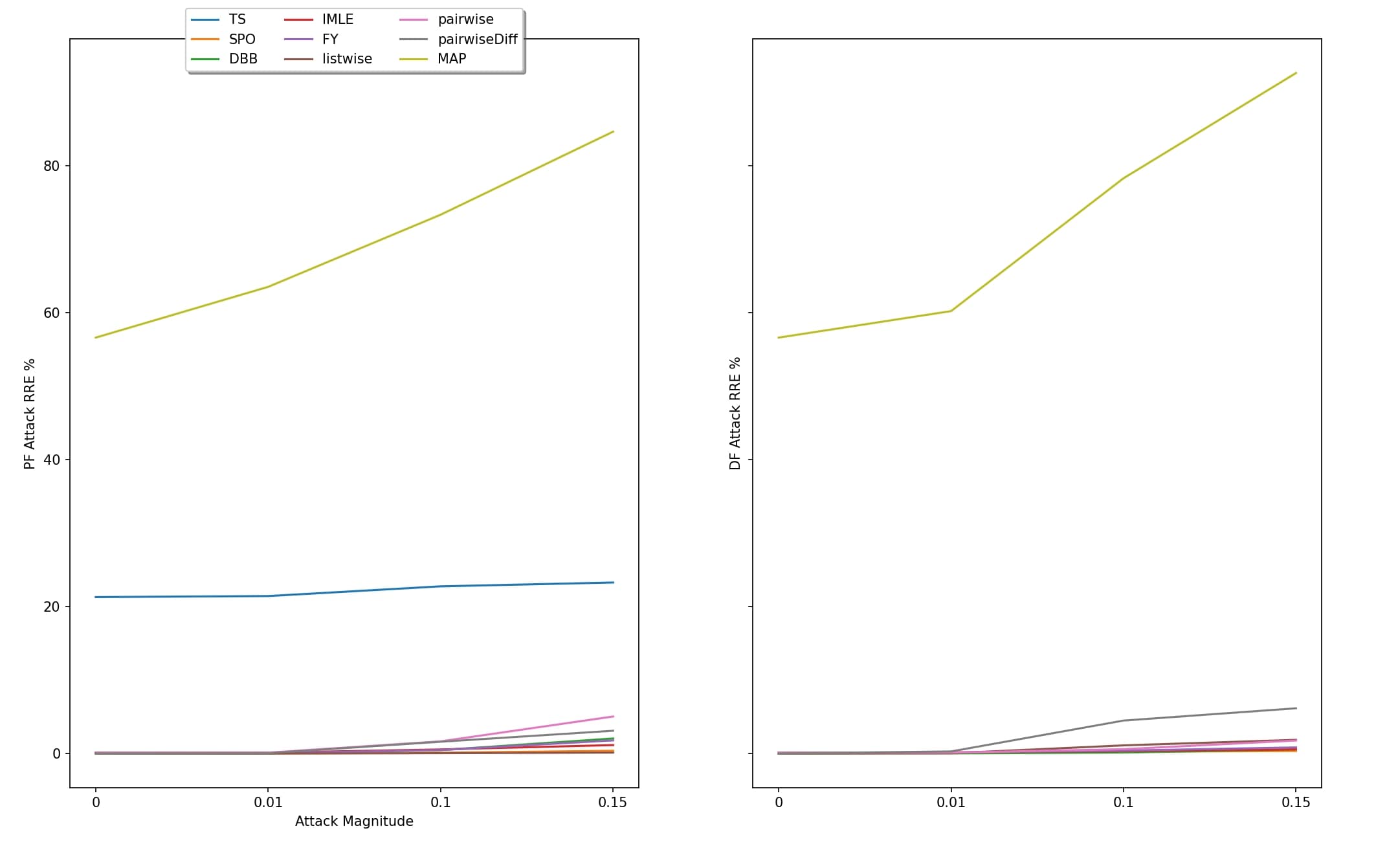} 
\caption{WarCraft 24 $\times$ 24 image size RRE, PF Attack left, DF attack right}
\label{War24_RRE}
\end{figure}

\section{Conclusions}
In this research, we propose a hypothesis centered around the decision quality robustness of DFL models: their robustness is intricately linked to the model's capacity to discern the optimal solution in alignment with the ground truth label. Our empirical evidence suggests a notable pattern: when models generate predictions that lead to optimal solutions but deviate from ground-truth labels, they tend to be more vulnerable to evasion attacks. Additionally, in examining the behavior of models that identify optimal solutions deviating from the ground-truth labels, we show how their specific optimality characteristics influence their susceptibility to differently targeted attacks. Our results reveal that such models respond differently to various attack strategies, each uniquely exploiting their particular form of optimality.

\appendix

\section{Synthetic Data generation Details}
\subsection{Portfolio Optimization}
The Synthetic input-target pairs $(\mathbf{z}, \mathbf{c})$ are randomly generated according to a random function with a specified degree
of nonlinearity $Deg \in \mathbb{N}$. The procedure for generating the data is as follows: 
For a number of assets $d$ and input features of size $p$. The input samples $x_{i} \in \mathbb{R}^p$ are independently and identically distributed, each element sampled from a standard normal distribution $N(0,1)$. A random matrix $B \in \mathbb{R}^{d \times p}$, whose elements $B_{ij} \in \{0,1\}$ are drown i.i.d. Bernoulli distributions with a probability of 0.5 for the value 1. For a selected noise level $\eta$, $L \in \mathbb{R}^{n \times 4}$ whose entries are drawn uniformly over $[-0.0025\eta, 0.0025\eta]$ is generated. Asset returns are calculated first in terms of their conditional mean as:
\[ \bar{c_{ij}} := (\frac{0.05}{\sqrt{p}}(B\mathbf{z_i})_{j}+(0.1)^{\frac{1}{Deg}})^{Deg} \]
The observed return vectors $\mathbf{c_i}$ are defined as $c_{ij} := \bar{r}_i + Lf + 0.01\eta\xi$, where $f \sim N(0,I_4)$ and noise $\xi \sim N(0, I_d)$. This results $c_{ij}$ to obey the covariance matrix $\Sigma := LL^{\top} + (0.01\xi)^2I$ which is used to form our problem constraint and bound the risk, which we define as $\lambda := 2.25e^{\top}\Sigma e$ where is a constant vector that represents the equal allocation solution. The value of the noise magnitude $\eta$ is set at 1. We assume that the covariance matrix  of the assest returns does not depend on the features. The values $\Sigma$ and $\lambda$ are constant and randomly generated for each setting.

\subsection{Knapsack}
The dataset, derived from the Irish Single Electricity Market Operator (SEMO) is structured such that each day represents an optimization case, and every half-hour is equivalent to an item in a knapsack problem. Accordingly, both the cost vector $c$ and weight vector $w$ have 48 elements, each representing a half-hour. Every cost vector element is linked to an 8-dimensional feature vector. The weight vector remains constant, and its values are synthetically generated using the same approach as in \cite{hardSOP} Each of the 48 half-hour periods is assigned a weight $w_i$ by choosing from the set {3, 5, 7}. To create a correlation between the weights of the items and their values, the energy price vector is multiplied by the weight vector. This is further randomized by adding Gaussian noise $\xi \sim N(0,25)$, resulting in the final item values $c_i$. The total weight for each instance is 240.

\section{Hyperparameter Selection}
The hyperparameter selection for each method and experiment are selected from \cite{mandi2023decisionfocused}. The authors find the best results through grid search. In this section we provide the list of the used hyperparameters for reproducibility. 

\begin{table*}[htbp]
\centering
\begin{tabular}{|l|c|c|}
\hline
\textbf{Capacity} & \textbf{60} & \textbf{120} \\ \hline
PF (lr) &  0.5 & 1. \\ \hline
SPO (lr) & 0.5 & 1. \\ \hline
DBB (lr, $\lambda$) & (0.5, 0.1) & (1., 1.) \\ \hline
IMLE (lr, $\lambda$, $\epsilon$, $\kappa$) & (0.5, 0.1, 0.5, 5) & (0.5, 0.1, 0.1, 5) \\ \hline
FY (lr, $\epsilon$) & (1., 0.005) & (1., 0.5) \\ \hline
IntOpt (lr, $\mu$, damping) & (0.5, 0.01, 10.) & (0.5, 0.1, 10.) \\ \hline
QPTL (lr, $\mu$) & (0.5, 10.) & (0.5, 1.) \\ \hline
Listwise (lr, $\tau$) & (1., 0.001) & (1., 0.001) \\ \hline
Pairwise (lr, $\Theta$) & (0.5, 10.) & (0.5, 10.) \\ \hline
PairwiseDiff (lr) & 1. & 1. \\ \hline
MAP (lr) & 1. & 1.\\ \hline
\end{tabular}
\caption{Optimal Hyperparameter Combination for Knapsack}
\label{knapHyper}
\end{table*}

\begin{table*}[htbp]
\centering
\begin{tabular}{|l|c|c|}
\hline
\textbf{Image Size} & \textbf{12} & \textbf{24} \\ \hline
PF (lr) & 0.001 & 0.001 \\ \hline
SPO (lr) & 0.005 & 0.005 \\ \hline
DBB (lr, $\lambda$) & (0.001, 10.) & (0.001, 100.) \\ \hline
IMLE (lr, $\lambda$, $\epsilon$, $\kappa$) & (0.001, 10., 0.05, 50) & (0.001, 10., 0.05, 50) \\ \hline
FY (lr, $\epsilon$) & (0.01, 0.01) & (0.01, 0.01) \\ \hline
Listwise (lr, $\tau$) & (0.005, 0.5) & (0.005, 0.5) \\ \hline
Pairwise (lr, $\Theta$) & (0.01, 0.1) & (0.01, 0.1) \\ \hline
PairwiseDiff (lr) & 0.005
 & 0.005
 \\ \hline
MAP (lr) & 0.005 & 0.005 \\ \hline
\end{tabular}
\caption{Optimal Hyperparameter Combination for Warcraft shortest path}
\label{WarcraftHyper}
\end{table*}

\begin{table*}[htbp]
\centering
\begin{tabular}{|l|c|c|}
\hline
\textbf{Deg} & \textbf{1} & \textbf{16} \\ \hline
PF (lr) & 0.01 & 0.05 \\ \hline
SPO (lr) & 0.5 & 0.5 \\ \hline
DBB (lr, $\lambda$) & (1., 0.1) & (1., 0.1) \\ \hline
IMLE (lr, $\lambda$, $\epsilon$, $\kappa$) & (0.5, 0.1, 0.1, 5) & (0.5, 0.1, 0.05,5) \\ \hline
FY (lr, $\epsilon$) & (0.1, 0.01) & (1., 2.) \\ \hline
QPTL (lr, $\mu$) & (0.1, 10.) & (0.05, 10.) \\ \hline
Listwise (lr, $\tau$) & (0.1, 0.01) & (0.05, 0.005) \\ \hline
Pairwise (lr, $\Theta$) & (0.01, 0.01) & (0.1, 0.05) \\ \hline
PairwiseDiff (lr) & 0.1 & 0.05 \\ \hline
MAP (lr) & 0.01 & 1. \\ \hline
\end{tabular}
\caption{Optimal Hyperparameter Combination for portfolio optimization}
\label{PortfolioHyper}
\end{table*}

\section{Appendix All Results}
In the main text, we present line plots for a single problem instance and the RRE metric to save space. This section includes all results and plots.


\begin{figure}[htbp]
\centering
\includegraphics[width = 8cm, height = 5cm]{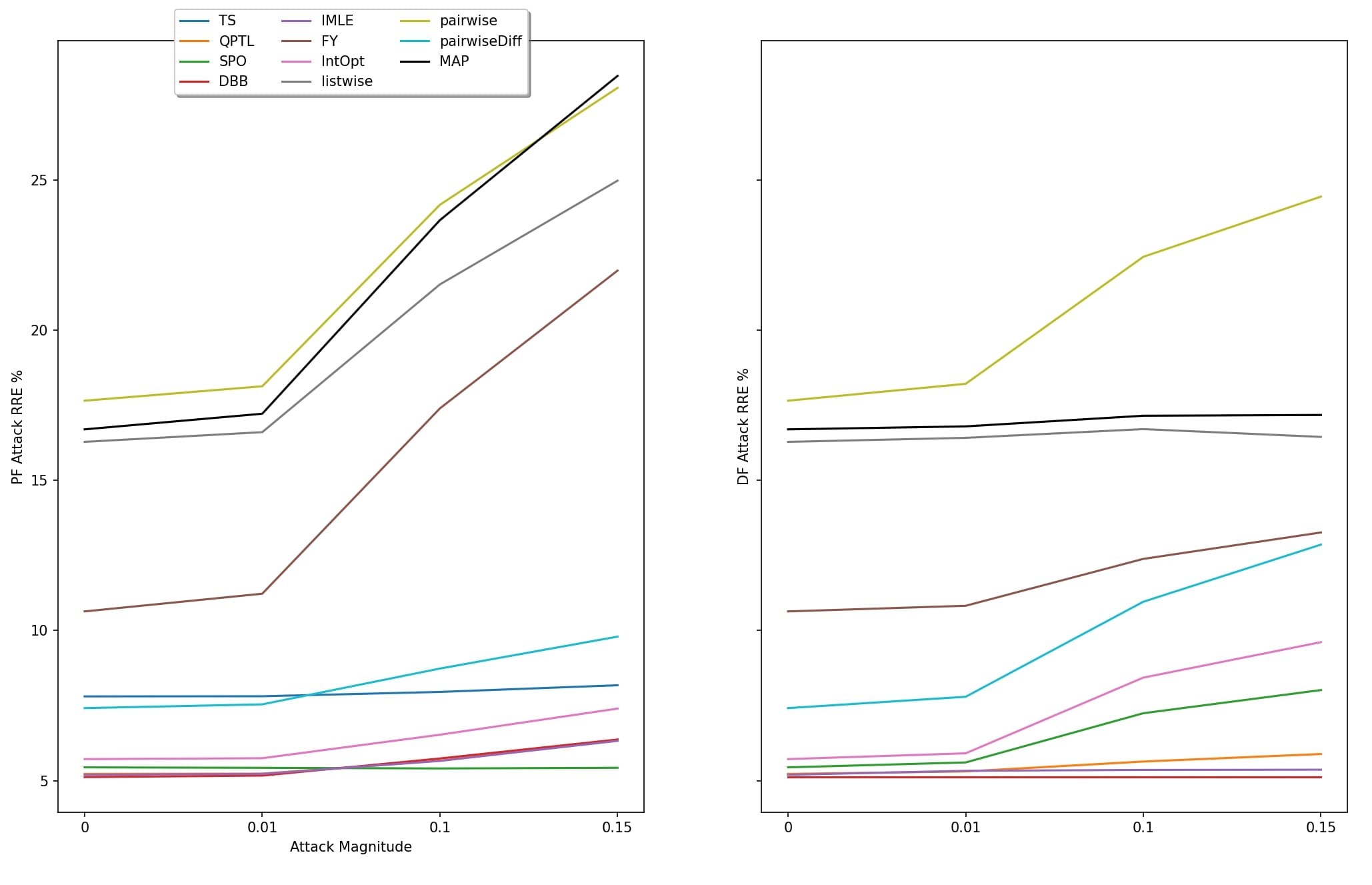} 
\caption{Knapsack capacity 120, RRE}
\label{AKnap120_RRE}
\end{figure}

\begin{figure}[htbp]
\centering
\includegraphics[width = 8cm, height = 5cm]{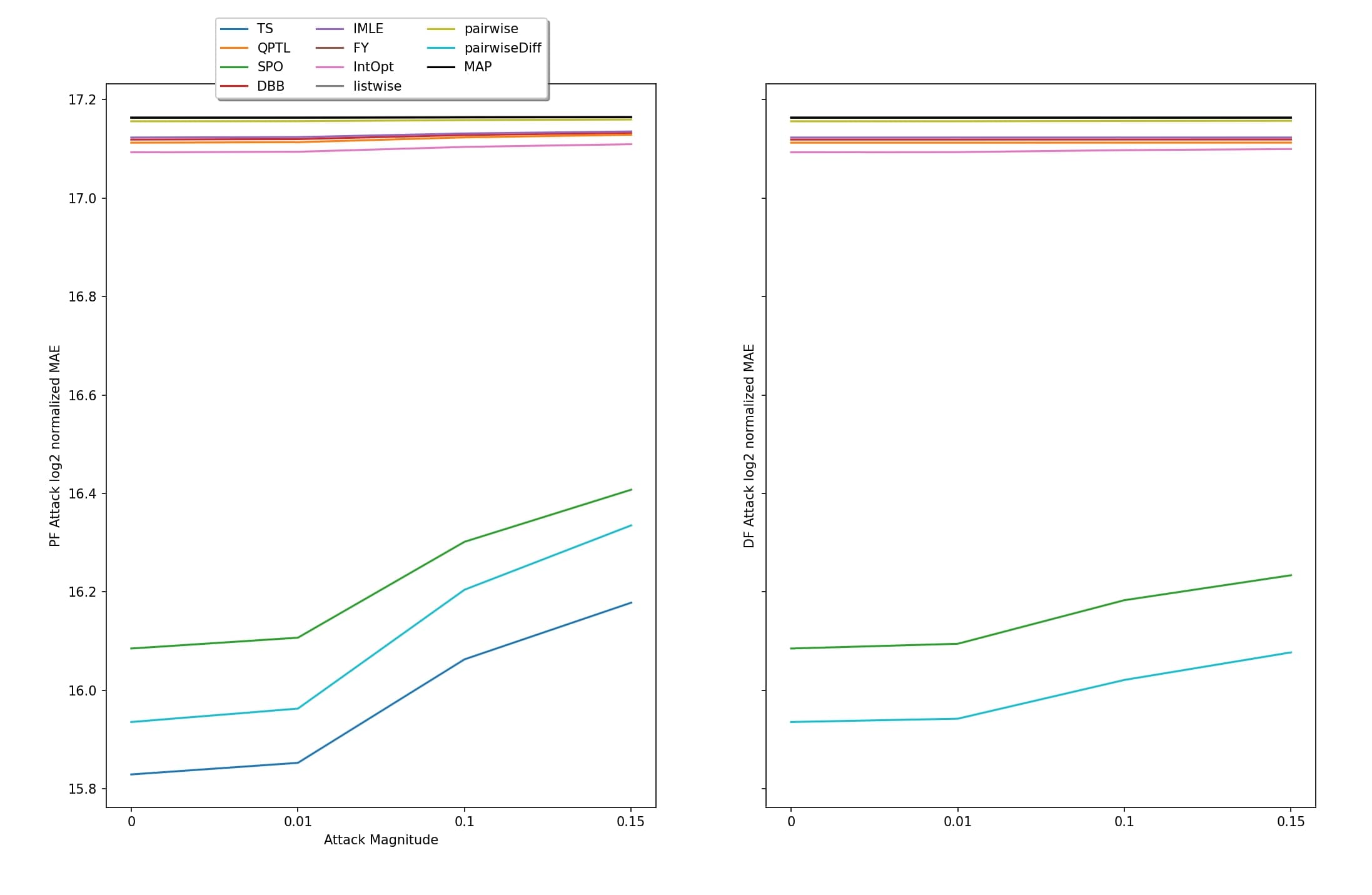} 
\caption{Knapsack capacity 120, MAE}
\label{AKnap120_MAE}
\end{figure}

\begin{figure}[htbp]
\centering
\includegraphics[width = 8cm, height = 5cm]{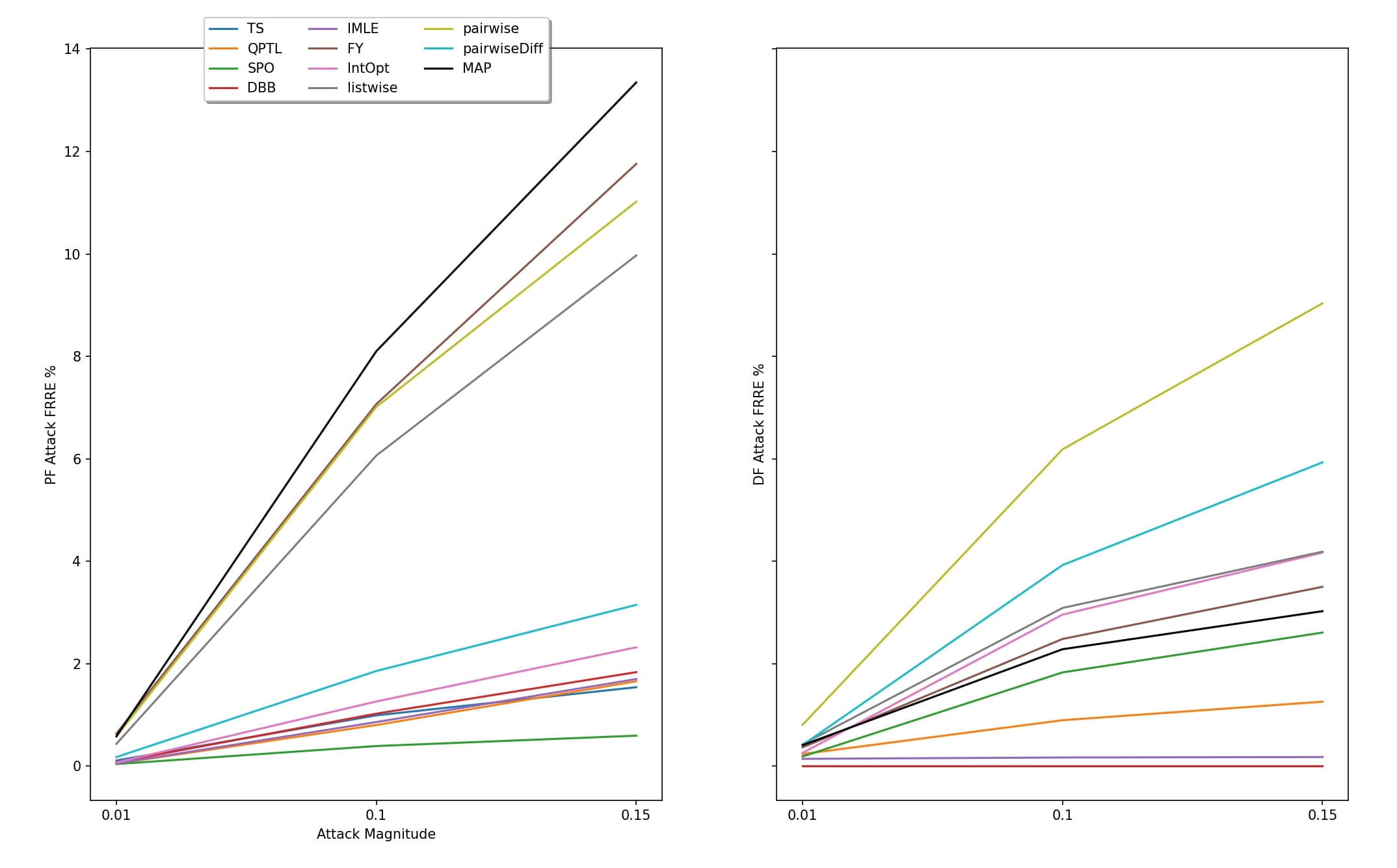} 
\caption{Knapsack capacity 120, FRRE}
\label{AKnap120_FRRE}
\end{figure}

\begin{figure}[htbp]
\centering
\includegraphics[width = 8cm, height = 5cm]{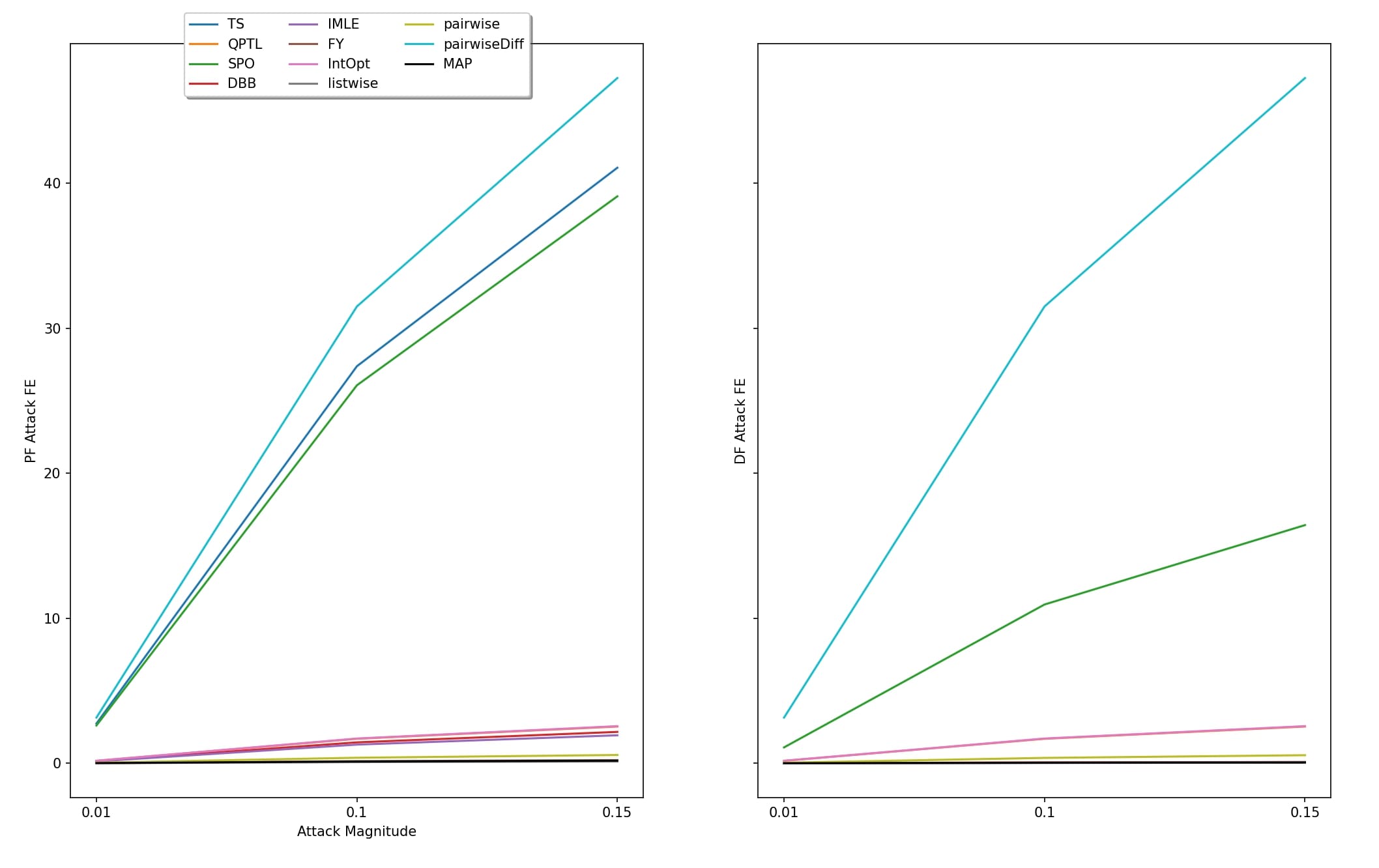} 
\caption{Knapsack capacity 120, FE}
\label{AKnap120_FE}
\end{figure}

\begin{figure}[htbp]
\centering
\includegraphics[width = 8cm, height = 5cm]{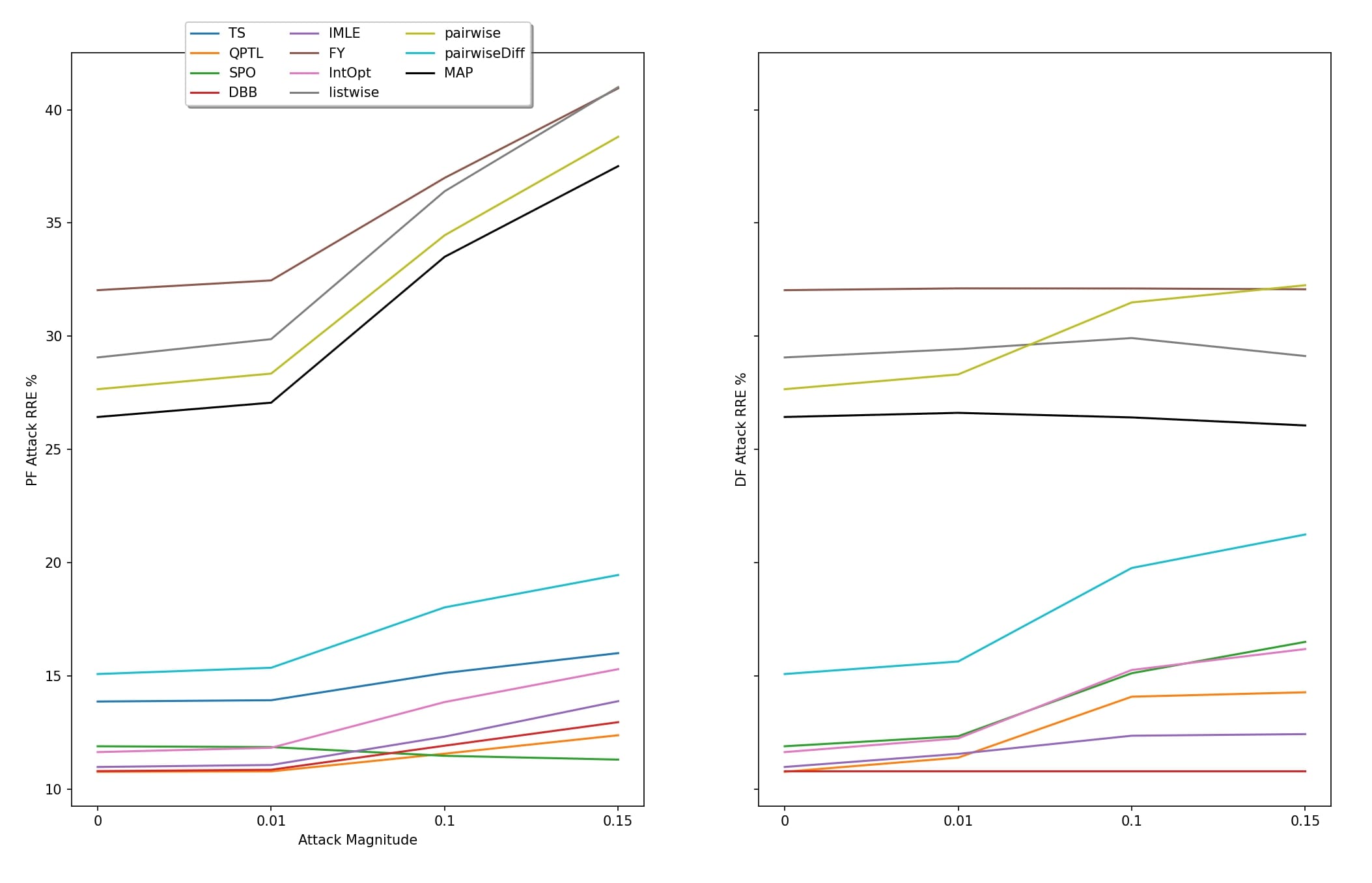} 
\caption{Knapsack capacity 60, RRE}
\label{AKnap60_RRE}
\end{figure}

\begin{figure}[htbp]
\centering
\includegraphics[width = 8cm, height = 5cm]{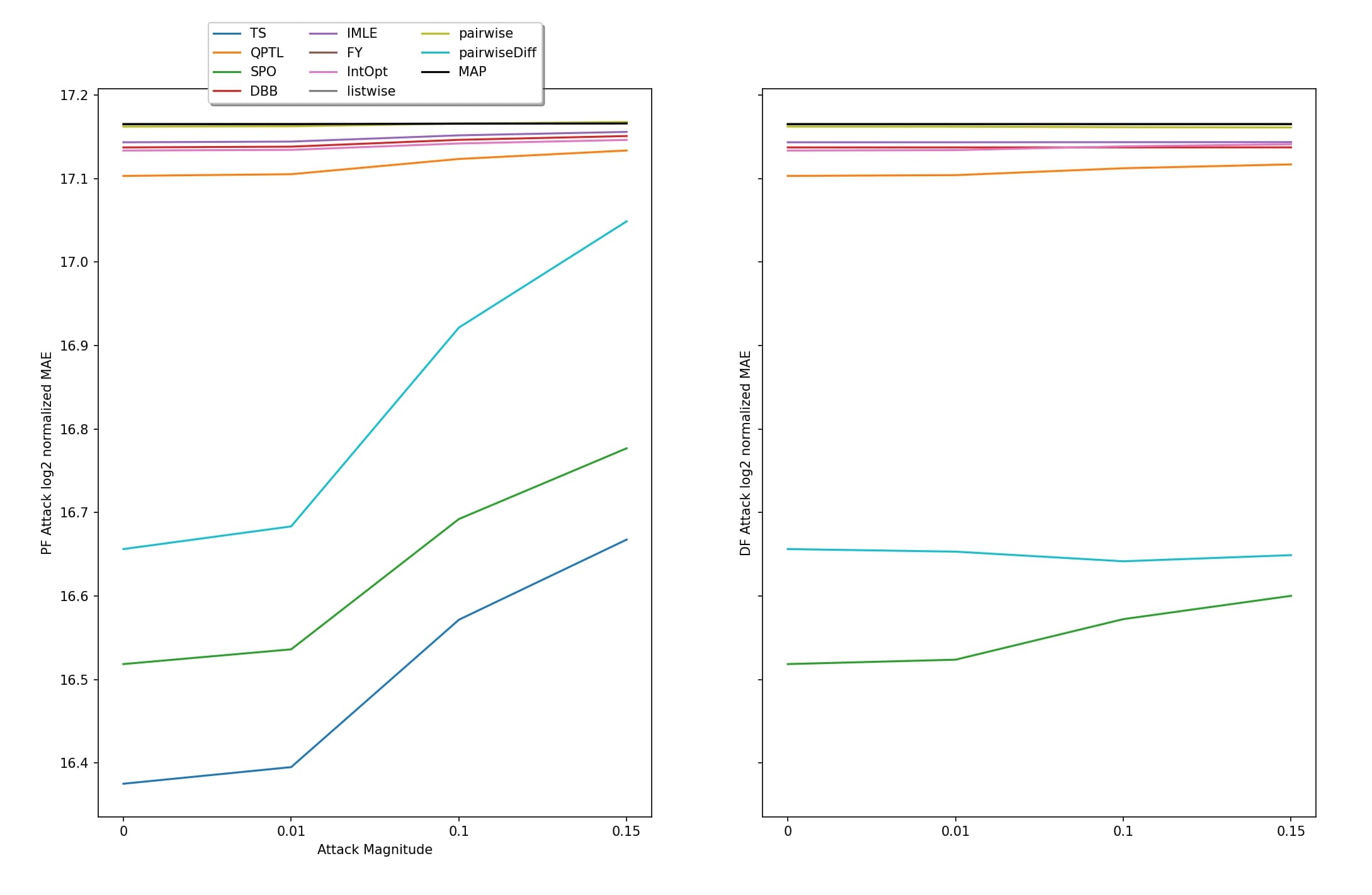} 
\caption{Knapsack capacity 60, MAE}
\label{AKnap60_MAE}
\end{figure}

\begin{figure}[htbp]
\centering
\includegraphics[width = 8cm, height = 5cm]{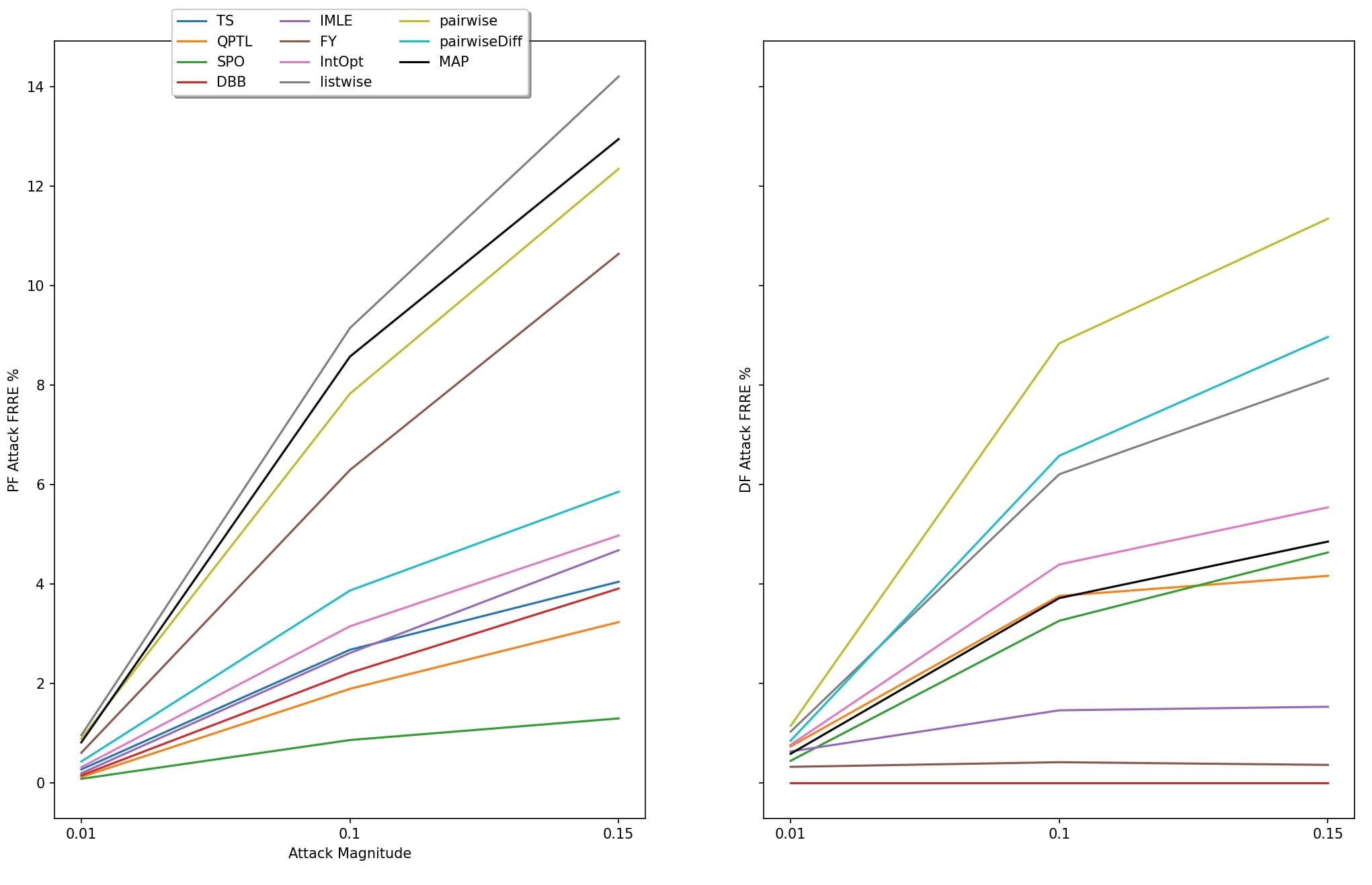} 
\caption{Knapsack capacity 60, FRRE}
\label{AKnap60_FRRE}
\end{figure}

\begin{figure}[htbp]
\centering
\includegraphics[width = 8cm, height = 5cm]{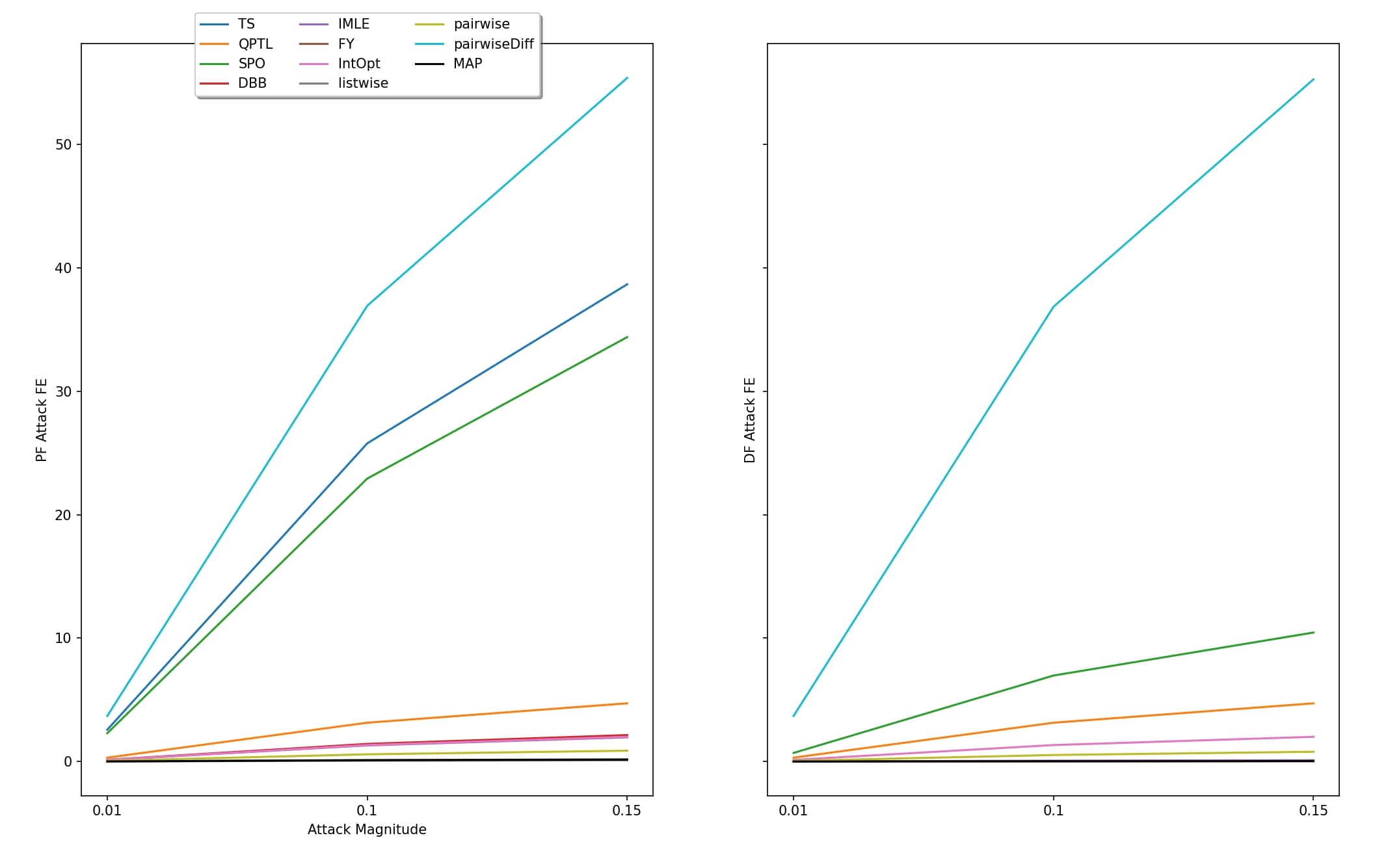} 
\caption{Knapsack capacity 60, FE}
\label{AKnap60_FE}
\end{figure}


\begin{figure}[htbp]
\centering
\includegraphics[width = 8cm, height = 5cm]{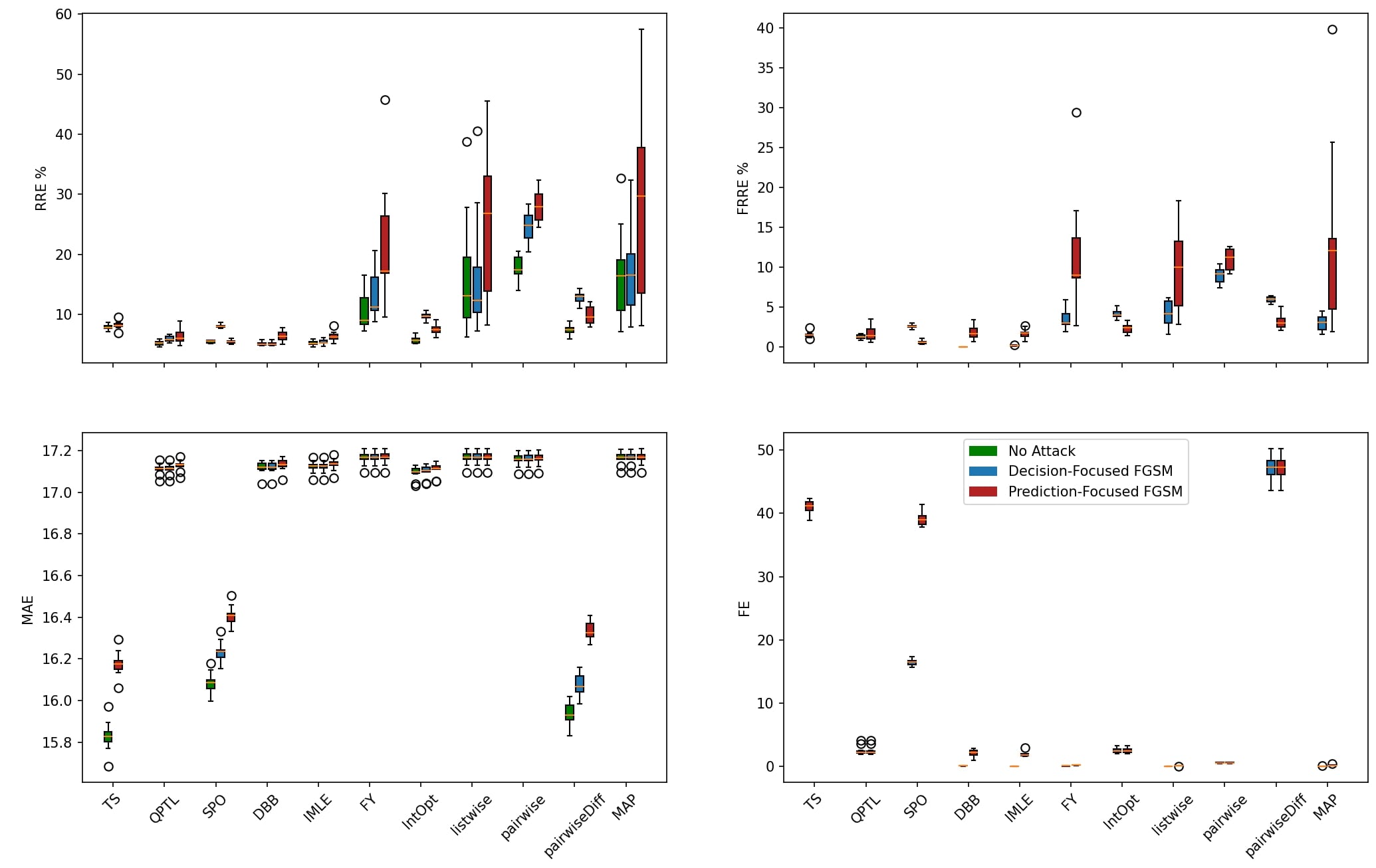} 
\caption{Knapsack capacity 120, Perturbation Magnitude 0.15}
\label{K120_0.15}
\end{figure}

\begin{figure}[htbp]
\centering
\includegraphics[width = 8cm, height = 5cm]{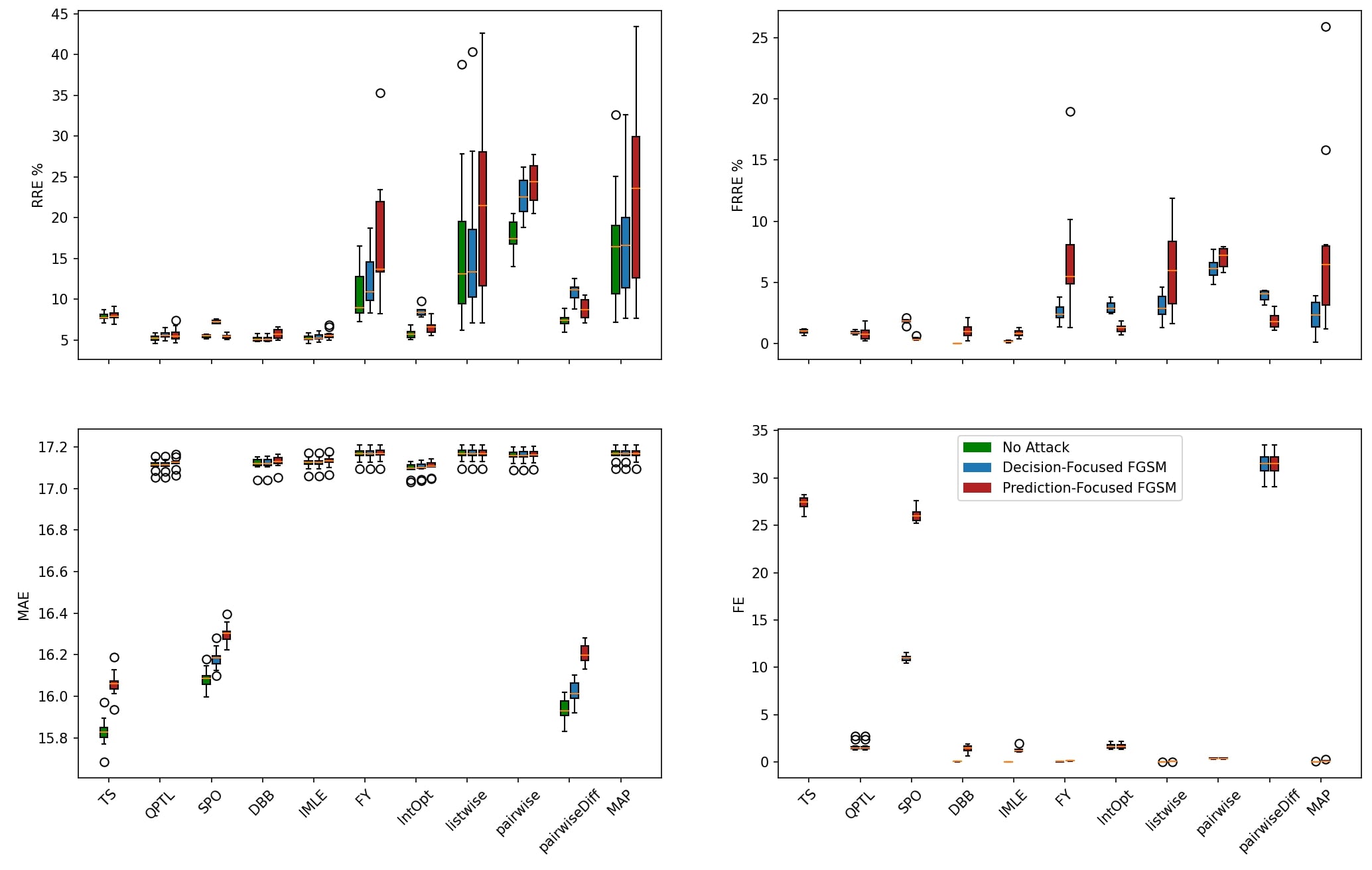} 
\caption{Knapsack capacity 120, Perturbation Magnitude 0.1}
\label{K120_0.1}
\end{figure}

\begin{figure}[htbp]
\centering
\includegraphics[width = 8cm, height = 5cm]{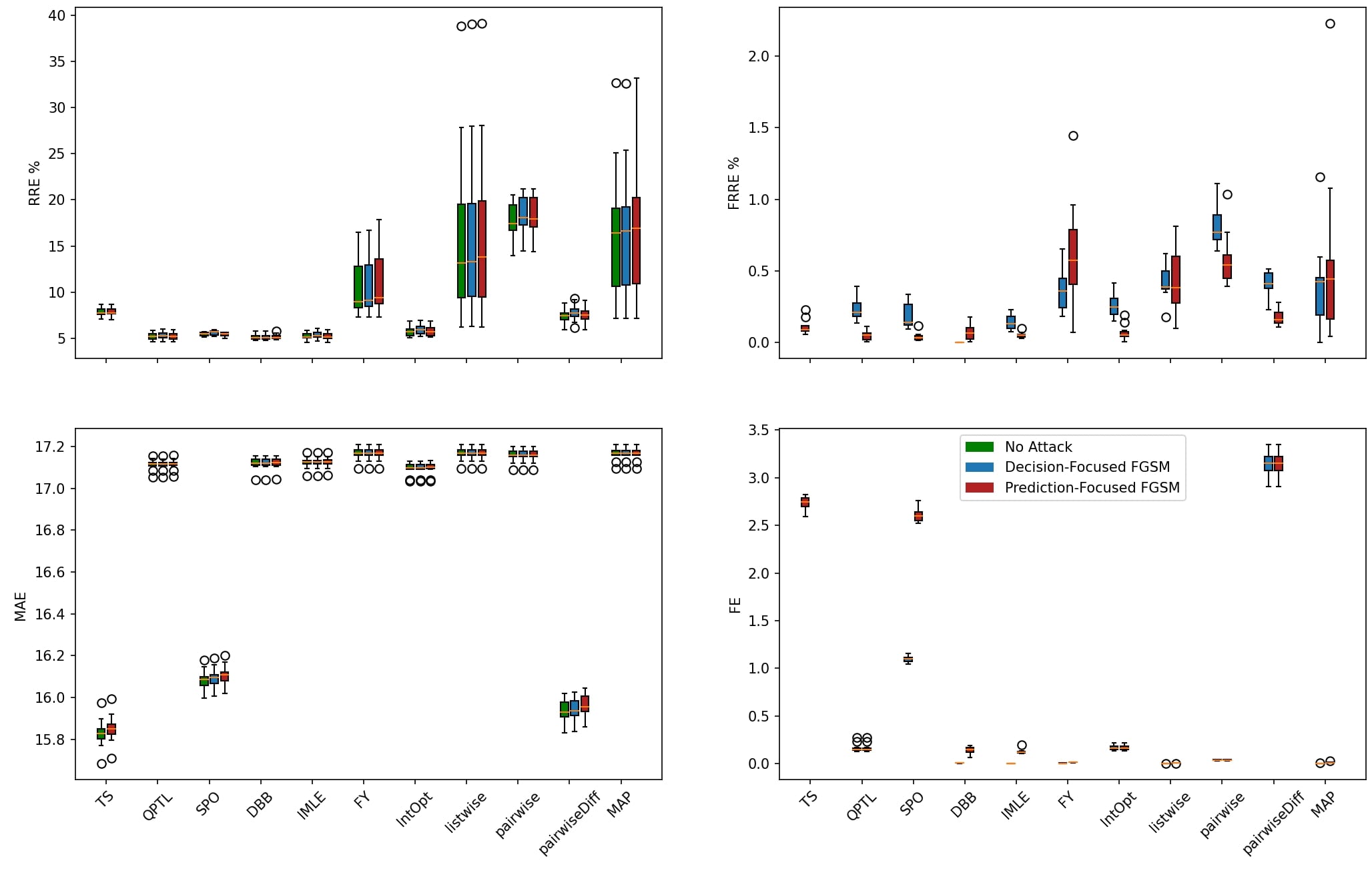} 
\caption{Knapsack capacity 120, Perturbation Magnitude 0.01}
\label{K120_0.01}
\end{figure}

\begin{figure}[htbp]
\centering
\includegraphics[width = 8cm, height = 5cm]{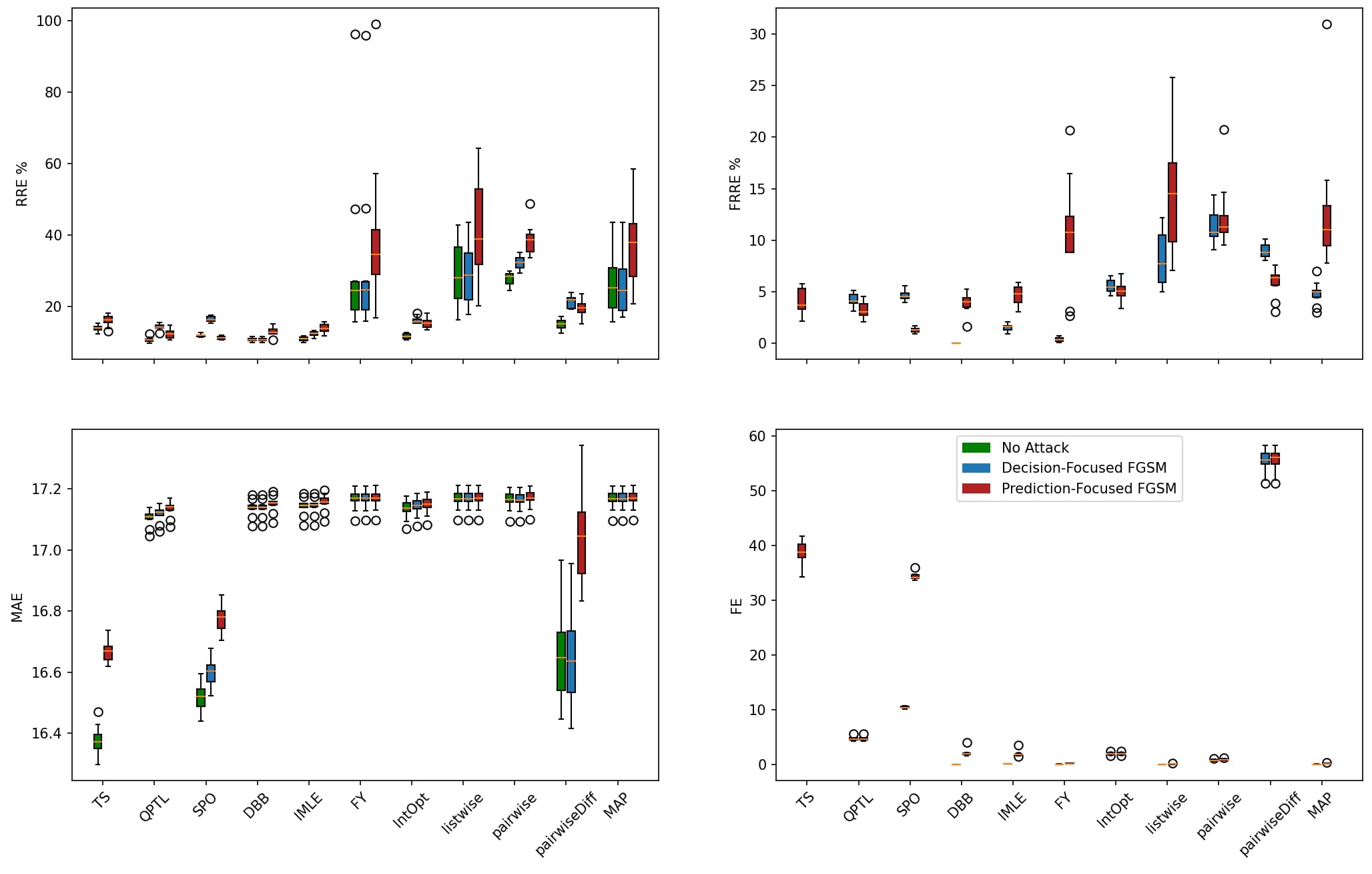} 
\caption{Knapsack capacity 60, Perturbation Magnitude 0.15}
\label{K60_0.15}
\end{figure}

\begin{figure}[htbp]
\centering
\includegraphics[width = 8cm, height = 5cm]{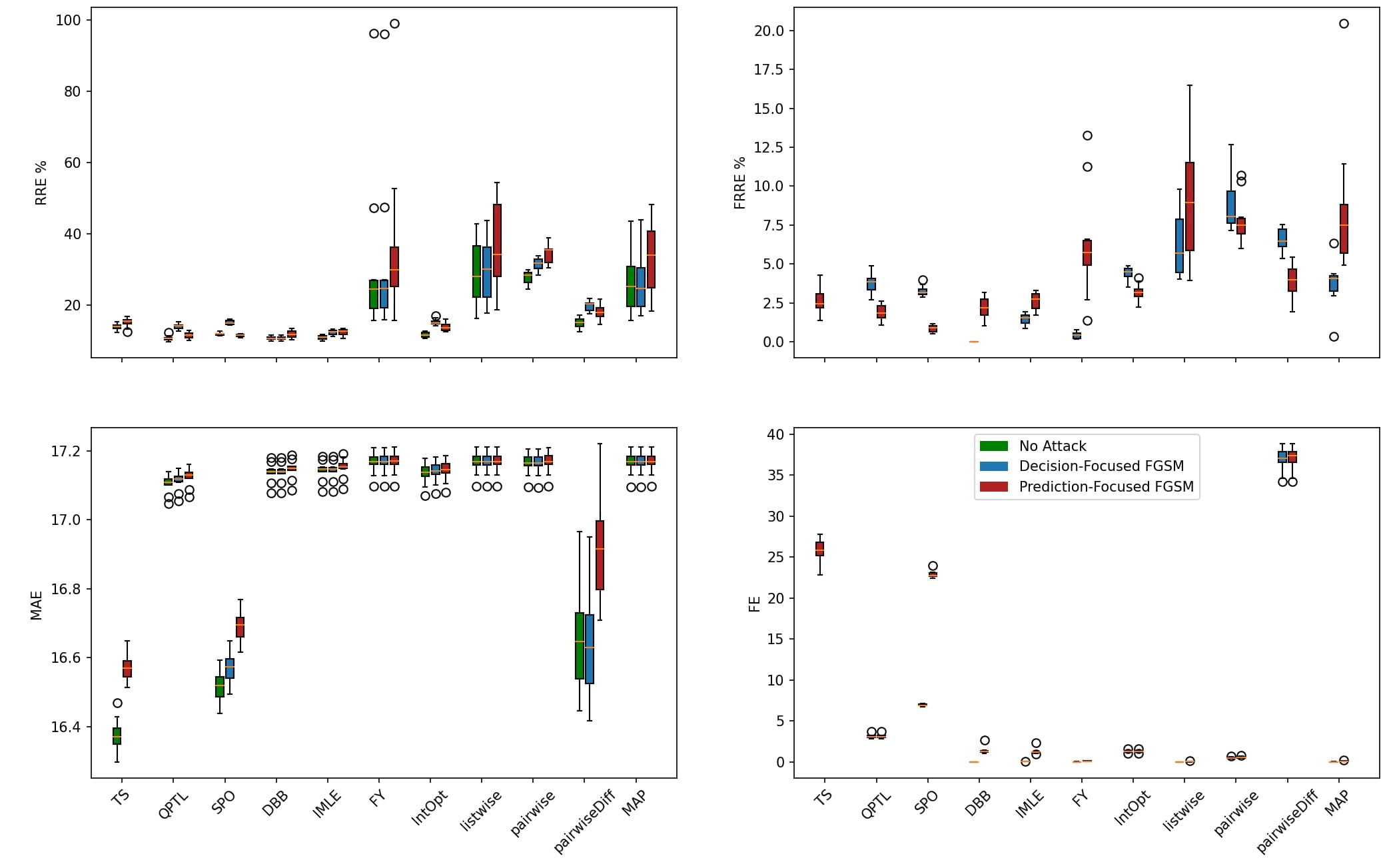} 
\caption{Knapsack capacity 60, Perturbation Magnitude 0.1}
\label{K60_0.1}
\end{figure}

\begin{figure}[htbp]
\centering
\includegraphics[width = 8cm, height = 5cm]{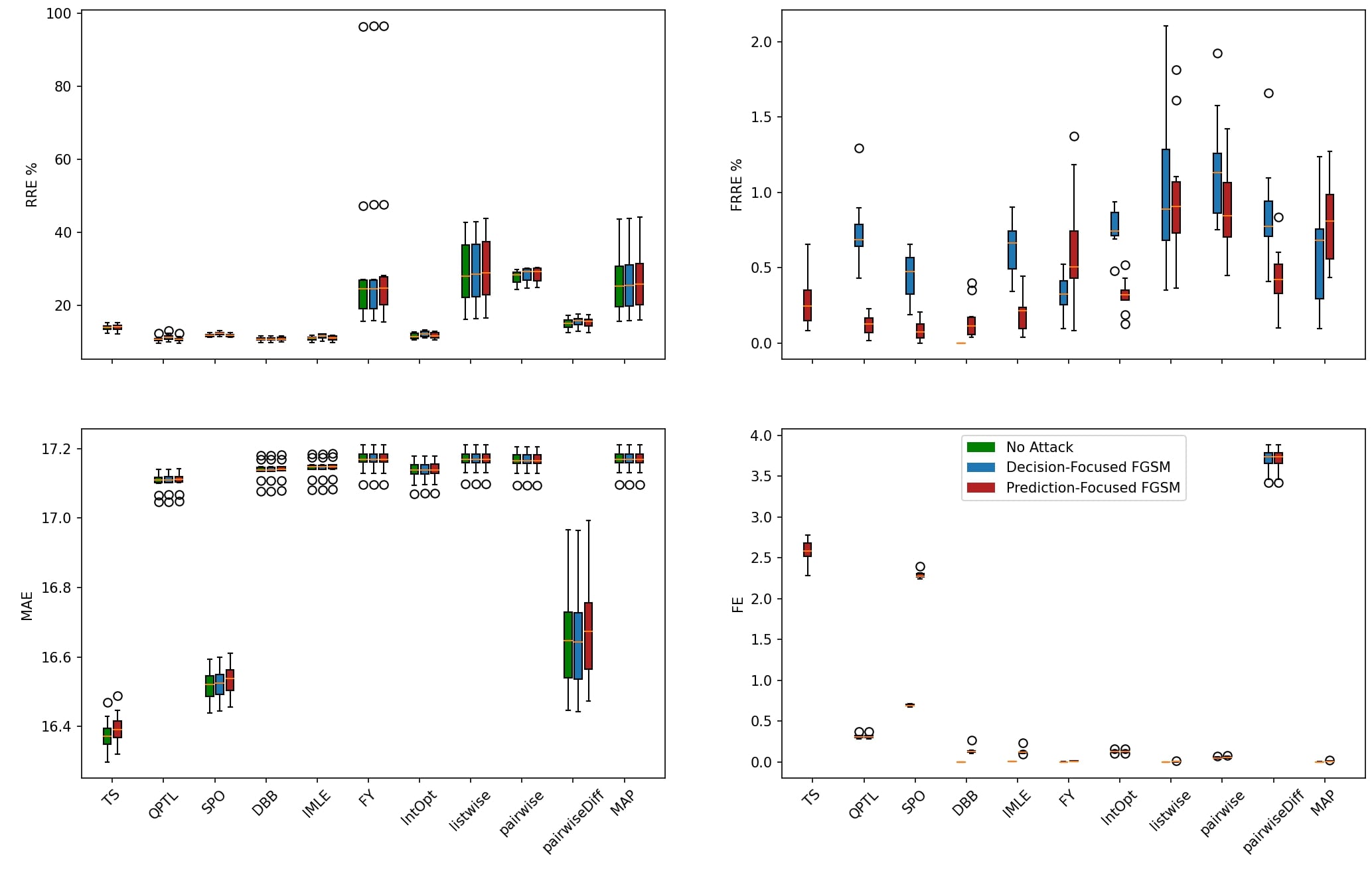} 
\caption{Knapsack capacity 60, Perturbation Magnitude 0.01}
\label{K60_0.01}
\end{figure}


\begin{figure}[htbp]
\centering
\includegraphics[width = 8cm, height = 5cm]{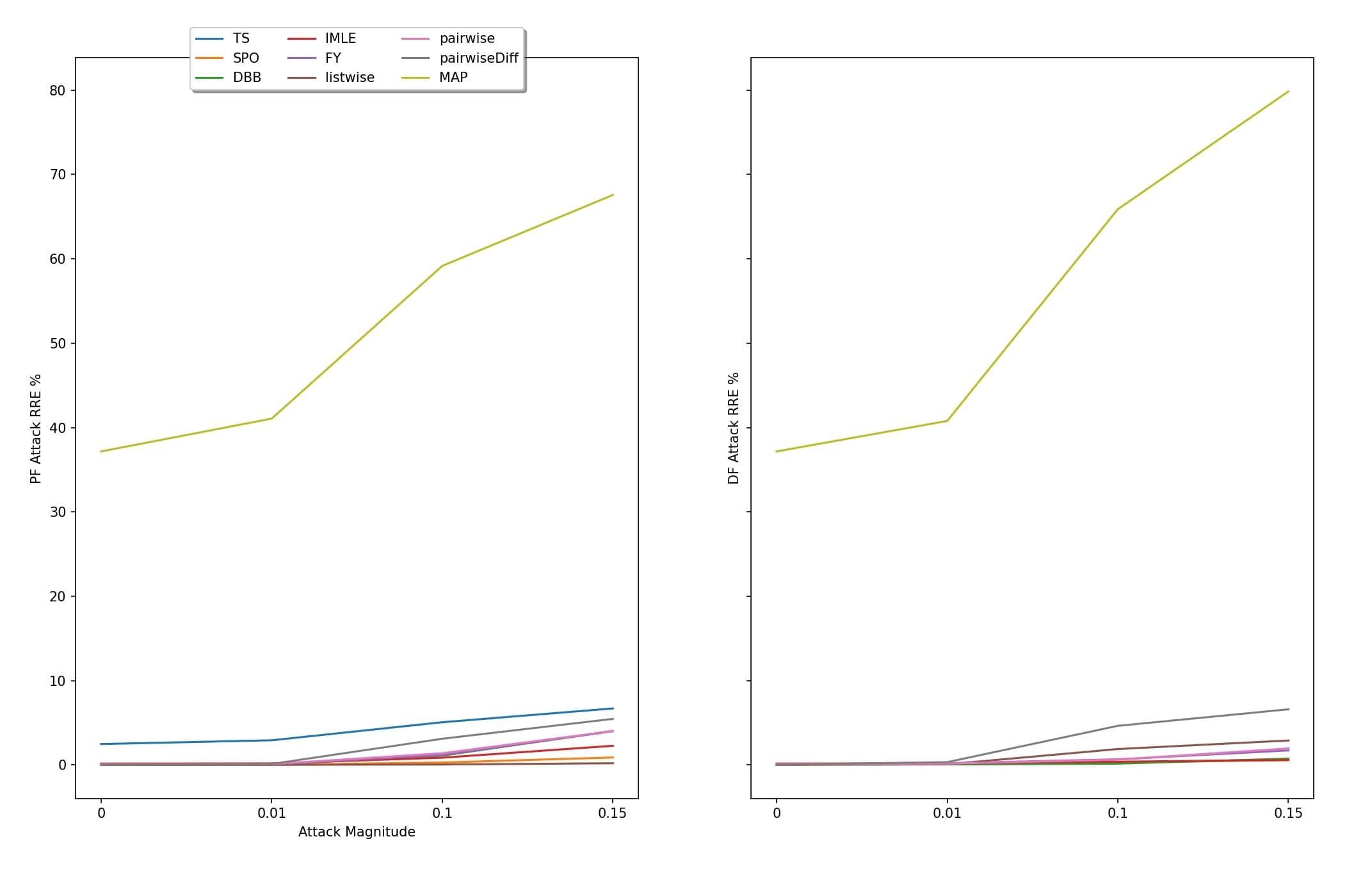} 
\caption{Warcraft img size $12 \times 12$, RRE }
\label{AW12_RRE}
\end{figure}

\begin{figure}[htbp]
\centering
\includegraphics[width = 8cm, height = 5cm]{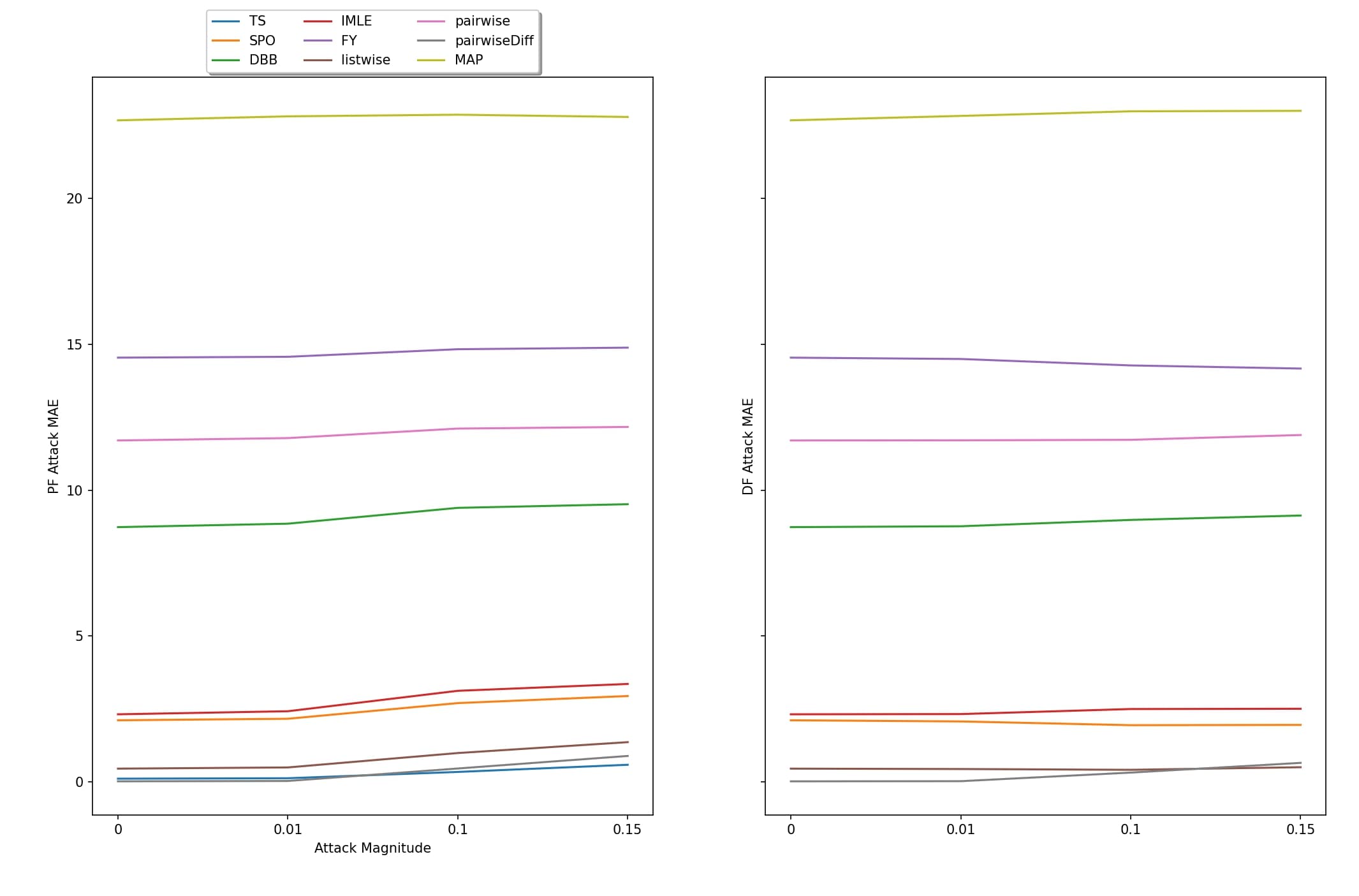} 
\caption{Warcraft img size $12 \times 12$,  MAE}
\label{AW12_MAE}
\end{figure}

\begin{figure}[htbp]
\centering
\includegraphics[width = 8cm, height = 5cm]{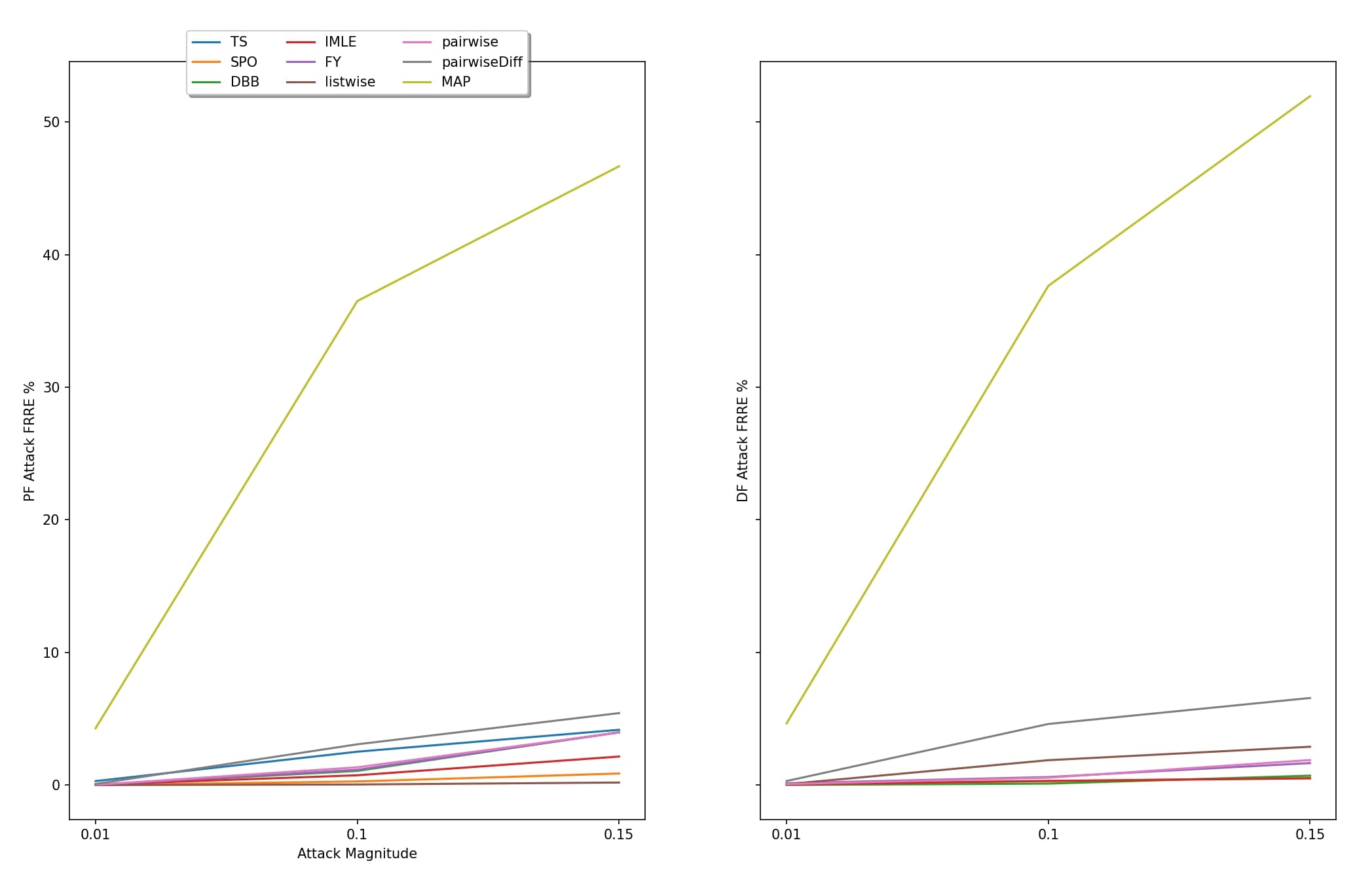} 
\caption{Warcraft img size $12 \times 12$, FRRE}
\label{AW12_FRRE}
\end{figure}

\begin{figure}[htbp]
\centering
\includegraphics[width = 8cm, height = 5cm]{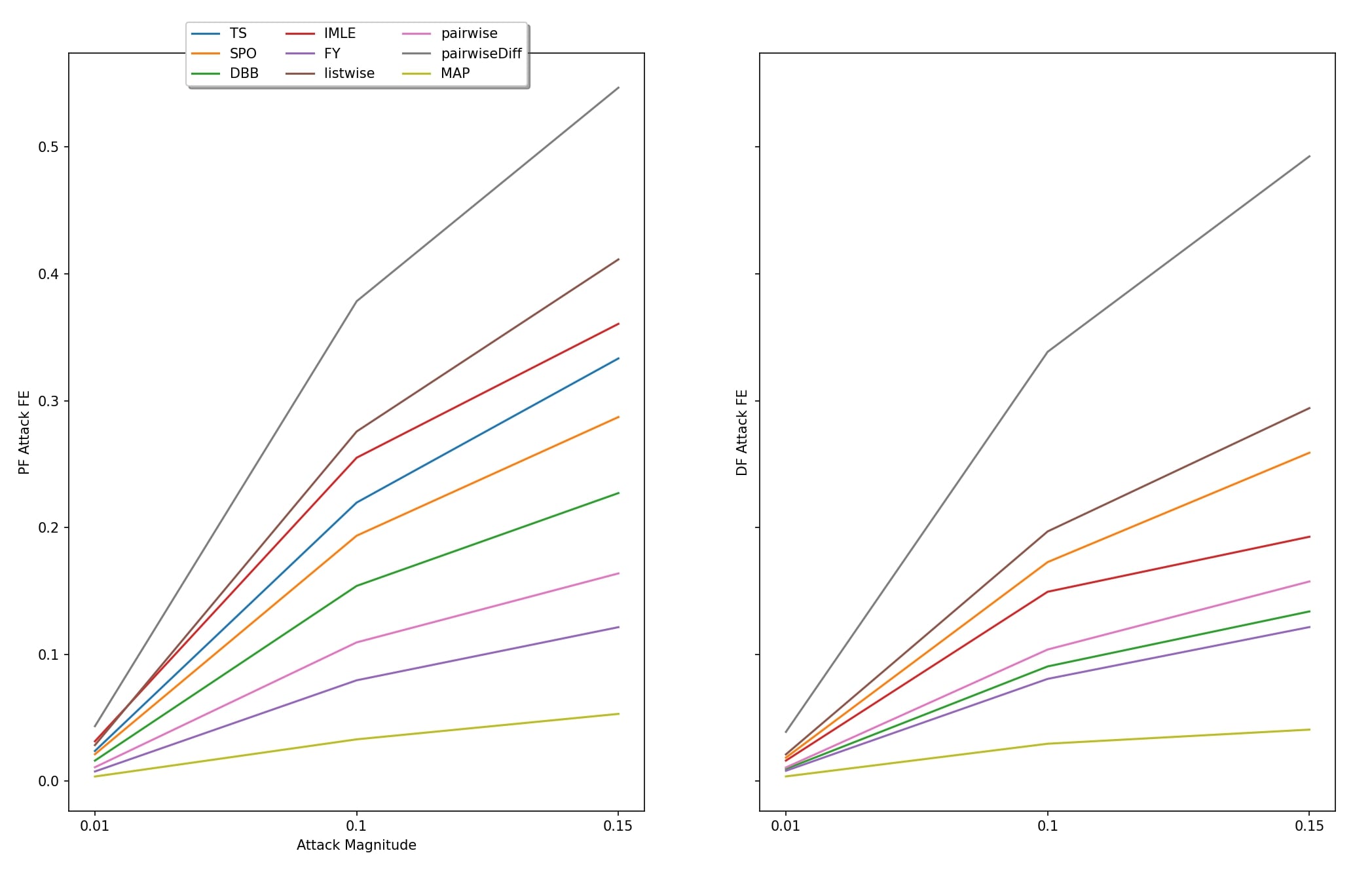} 
\caption{Warcraft img size $12 \times 12$, FE}
\label{AW12_FE}
\end{figure}

\begin{figure}[htbp]
\centering
\includegraphics[width = 8cm, height = 5cm]{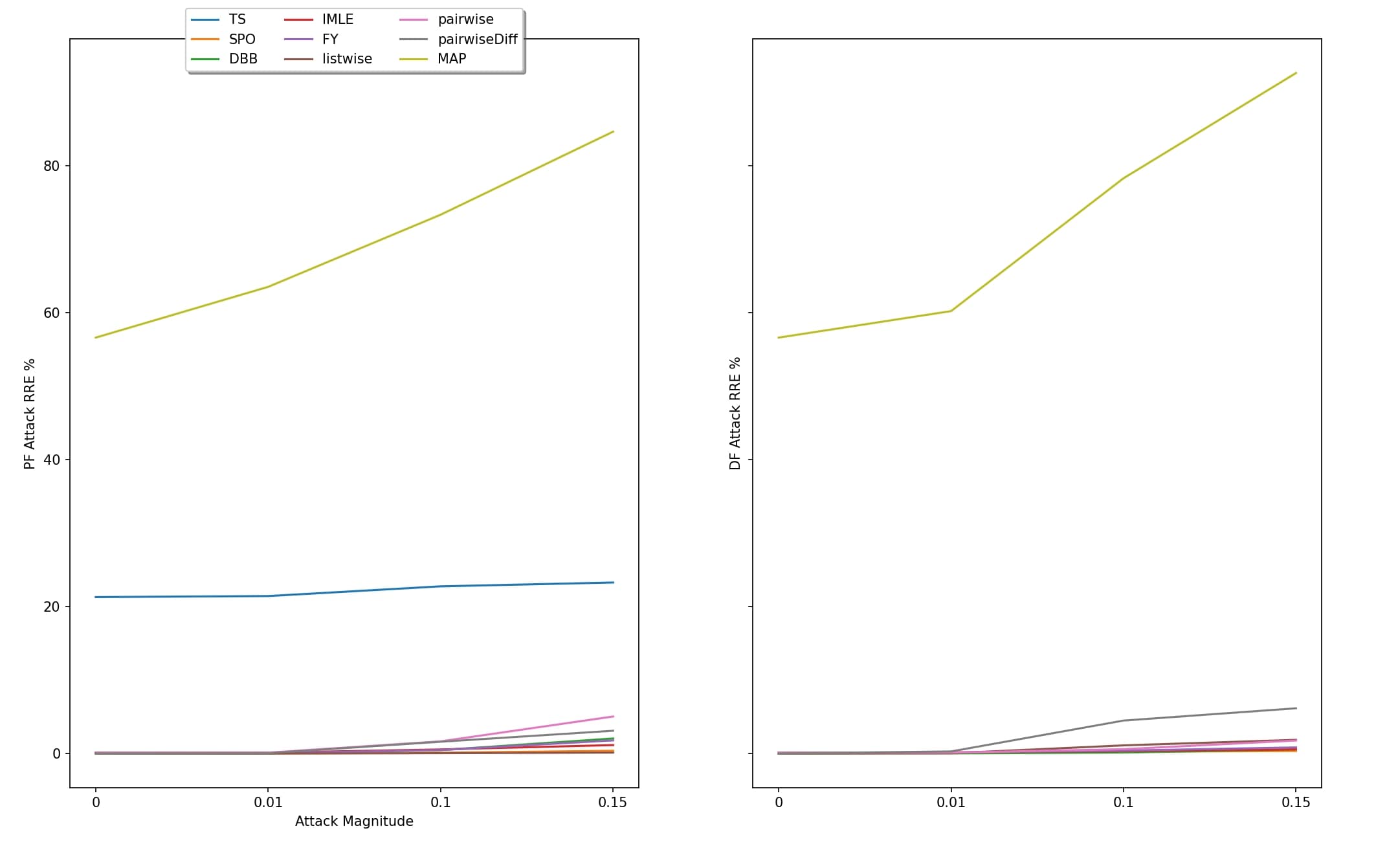} 
\caption{Warcraft img size $24 \times 24$, RRE}
\label{AW24_RRE}
\end{figure}

\begin{figure}[htbp]
\centering
\includegraphics[width = 8cm, height = 5cm]{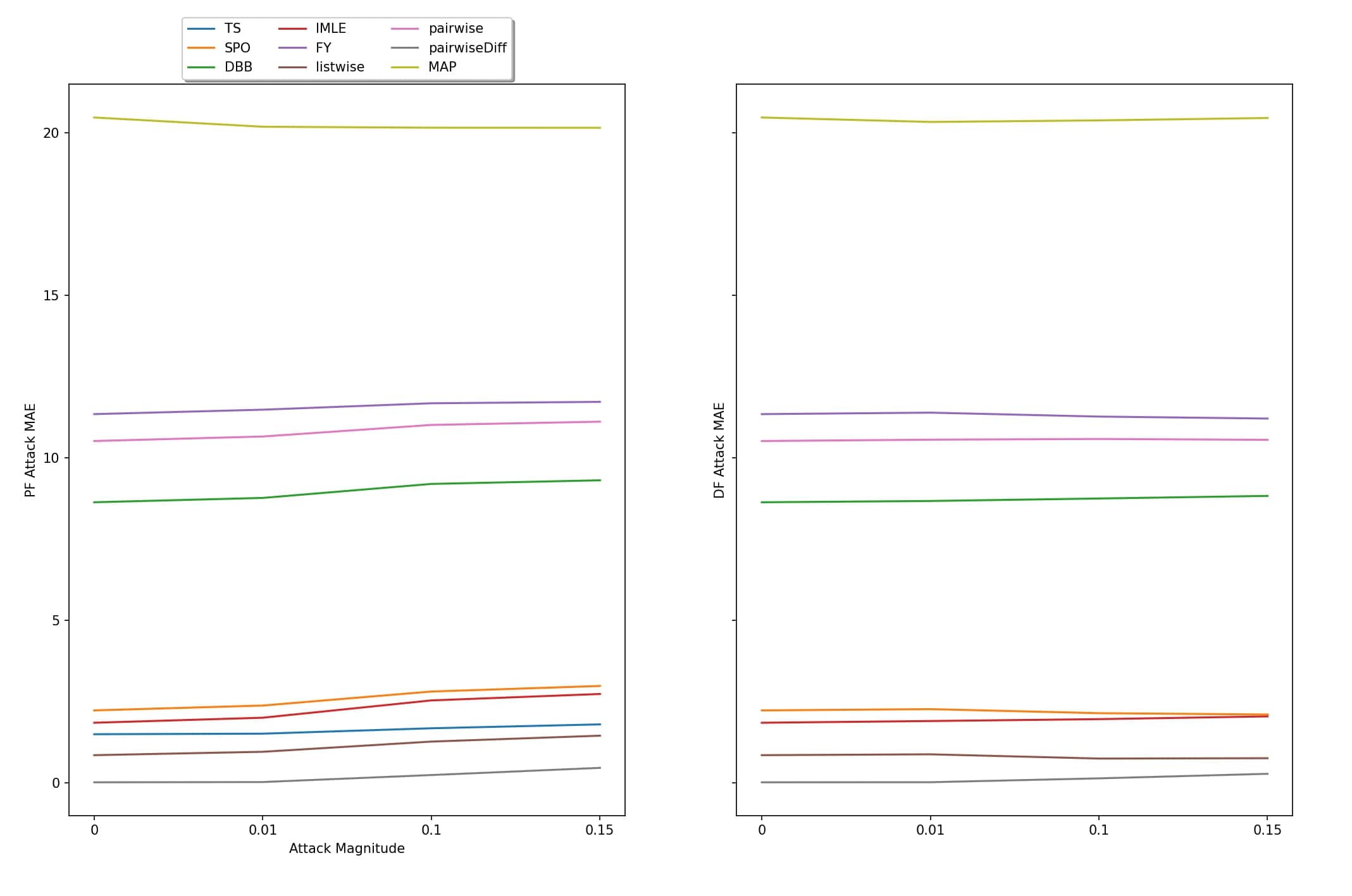} 
\caption{Warcraft img size $24 \times 24$,  MAE}
\label{AW24_MAE}
\end{figure}

\begin{figure}[htbp]
\centering
\includegraphics[width = 8cm, height = 5cm]{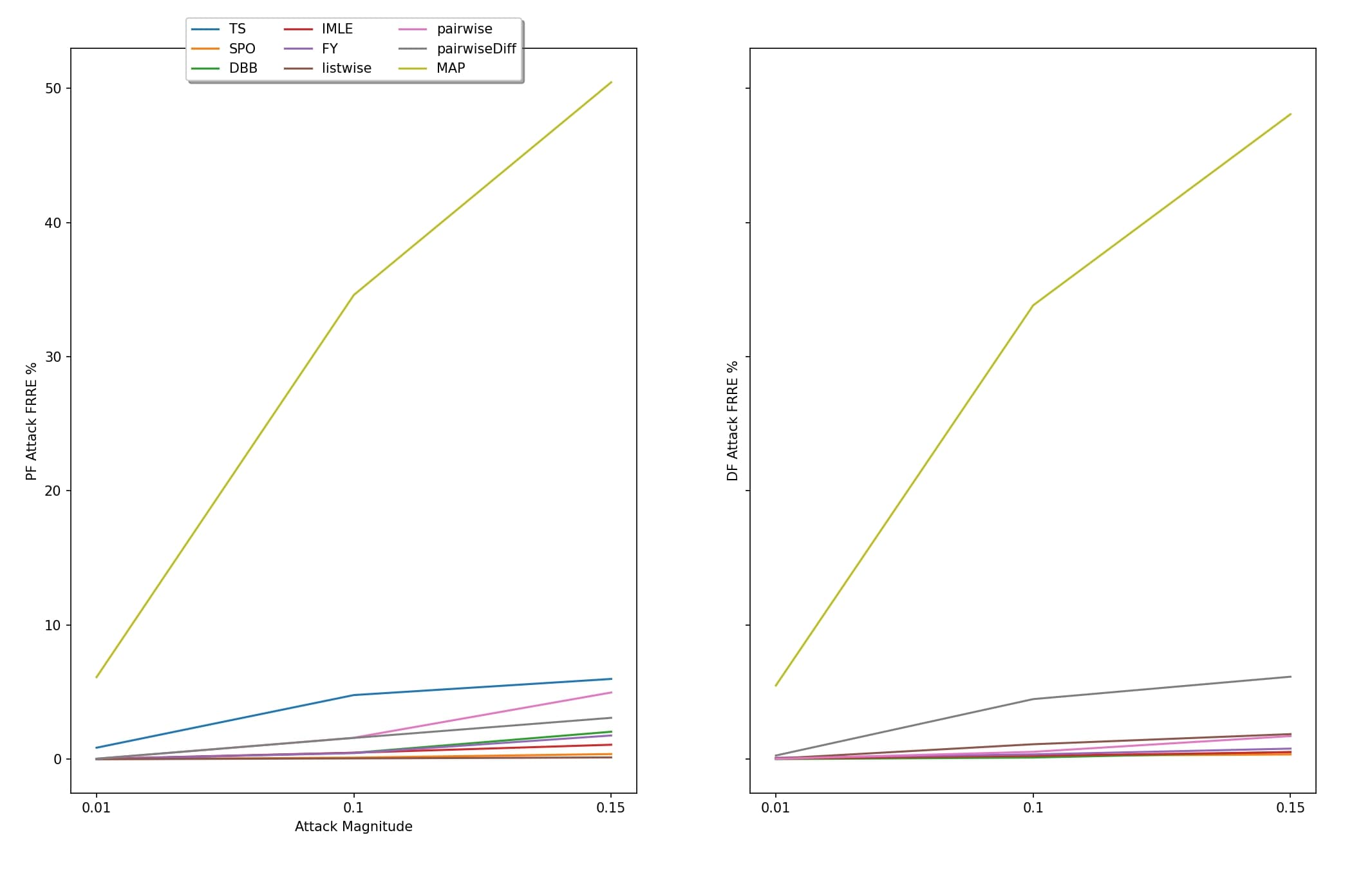} 
\caption{Warcraft img size $24 \times 24$, FRRE}
\label{AW24_FRRE}
\end{figure}

\begin{figure}[htbp]
\centering
\includegraphics[width = 8cm, height = 5cm]{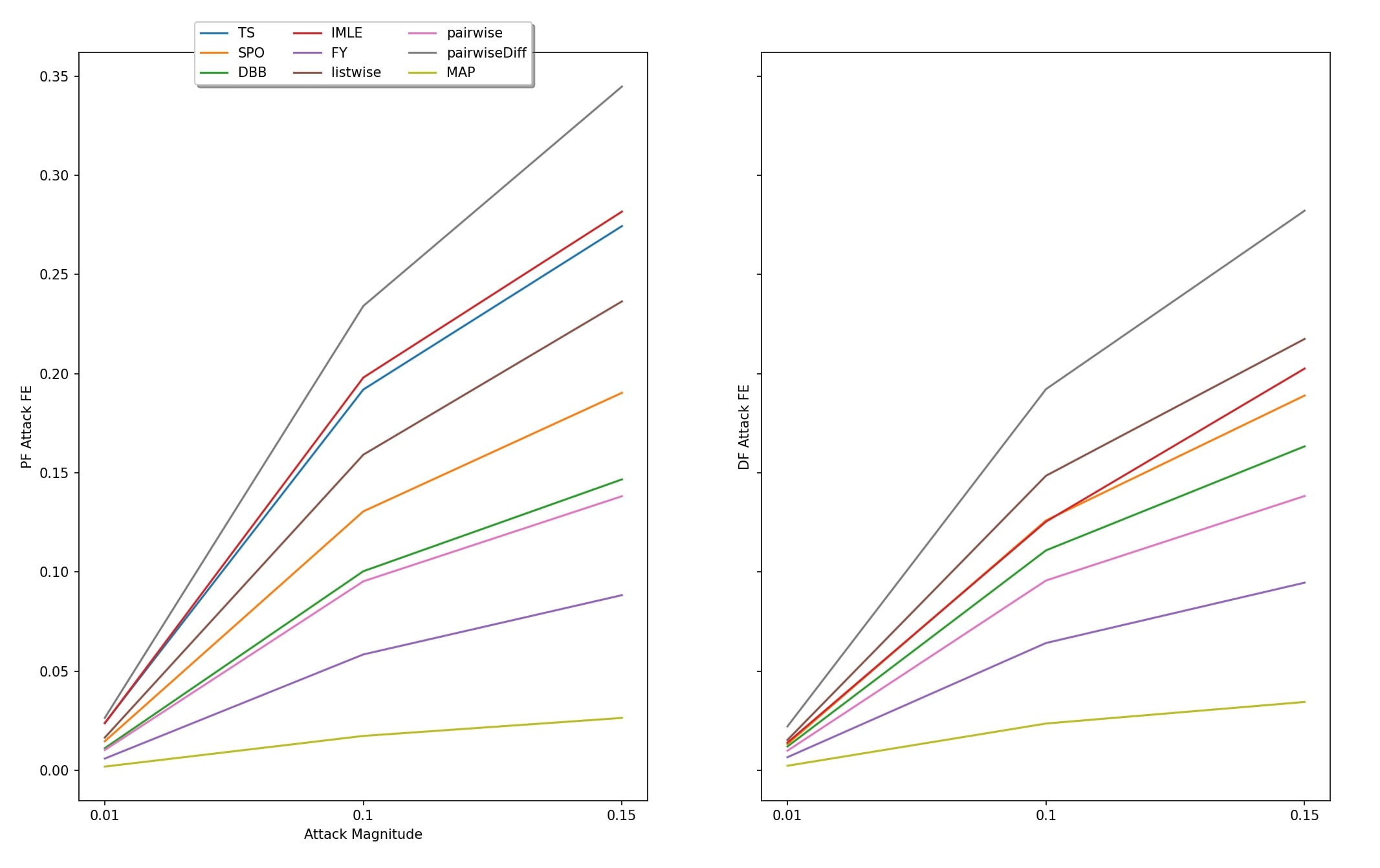} 
\caption{Warcraft img size $24 \times 24$, FE}
\label{AW24_FE}
\end{figure}


\begin{figure}[htbp]
\centering
\includegraphics[width = 8cm, height = 5cm]{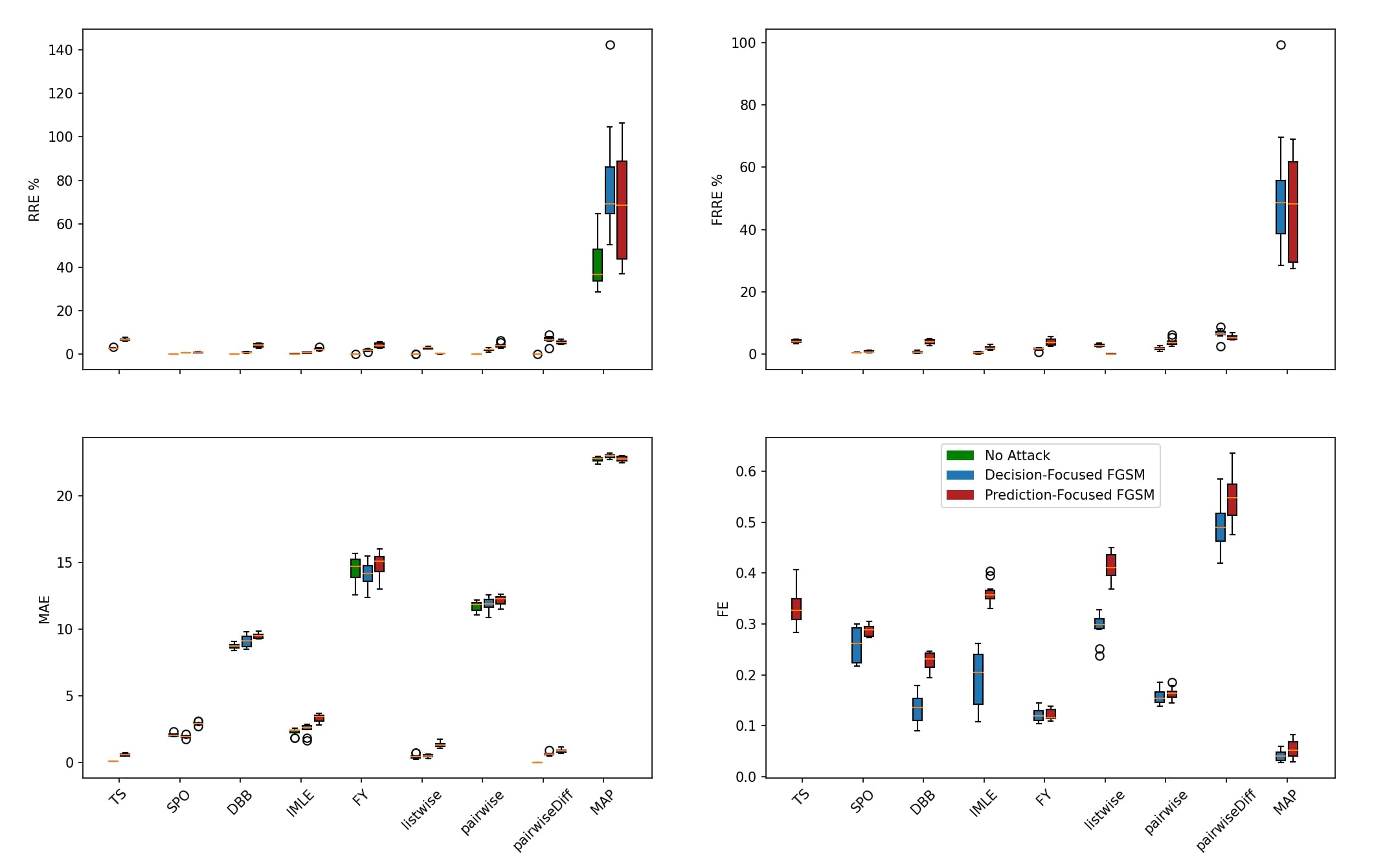} 
\caption{WarCraft img size $12 \times 12$, Perturbation Magnitude 0.15}
\label{W12_0.15}
\end{figure}

\begin{figure}[htbp]
\centering
\includegraphics[width = 8cm, height = 5cm]{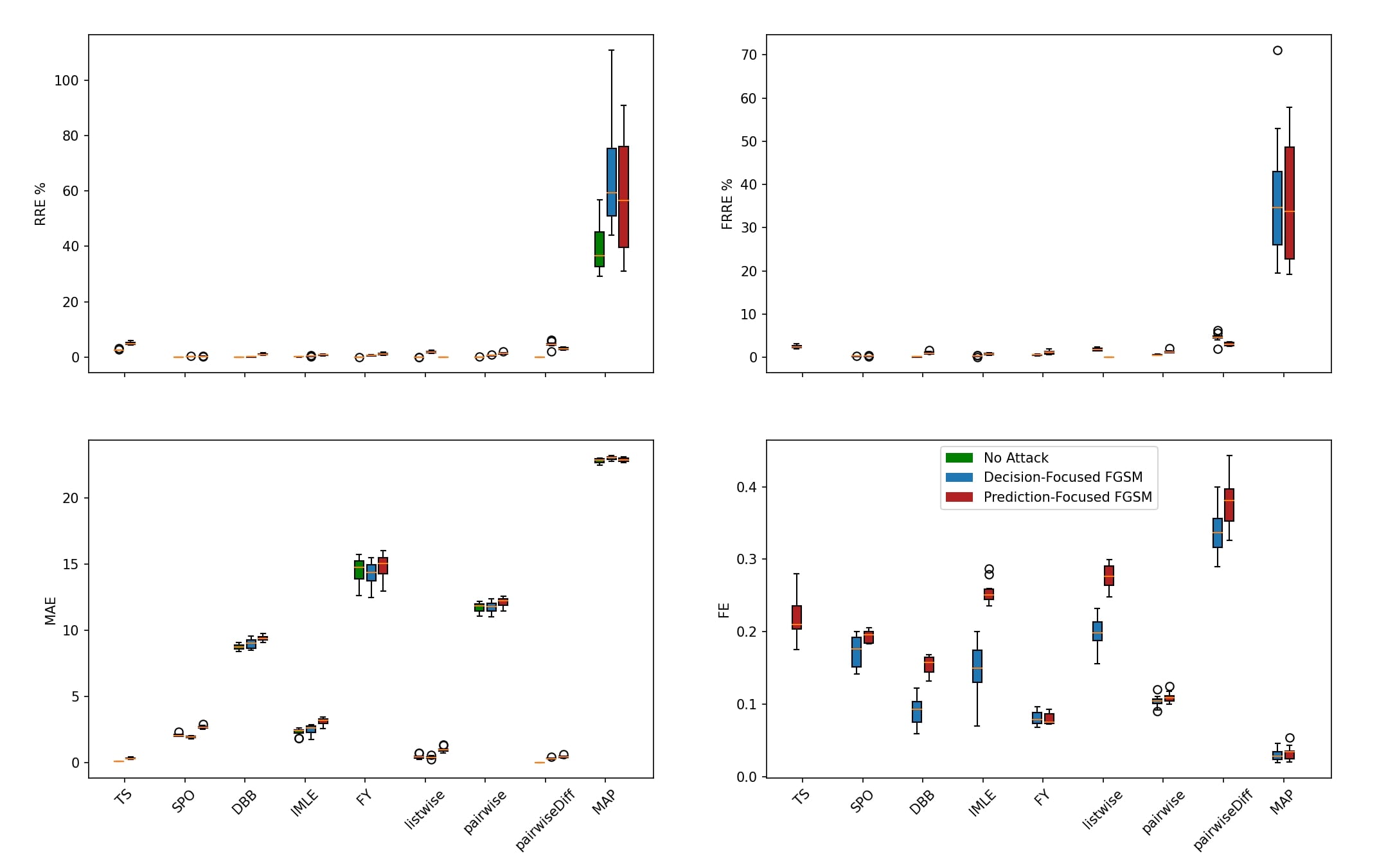} 
\caption{WarCraft img size $12 \times 12$, Perturbation Magnitude 0.1}
\label{W12_0.1}
\end{figure}

\begin{figure}[htbp]
\centering
\includegraphics[width = 8cm, height = 5cm]{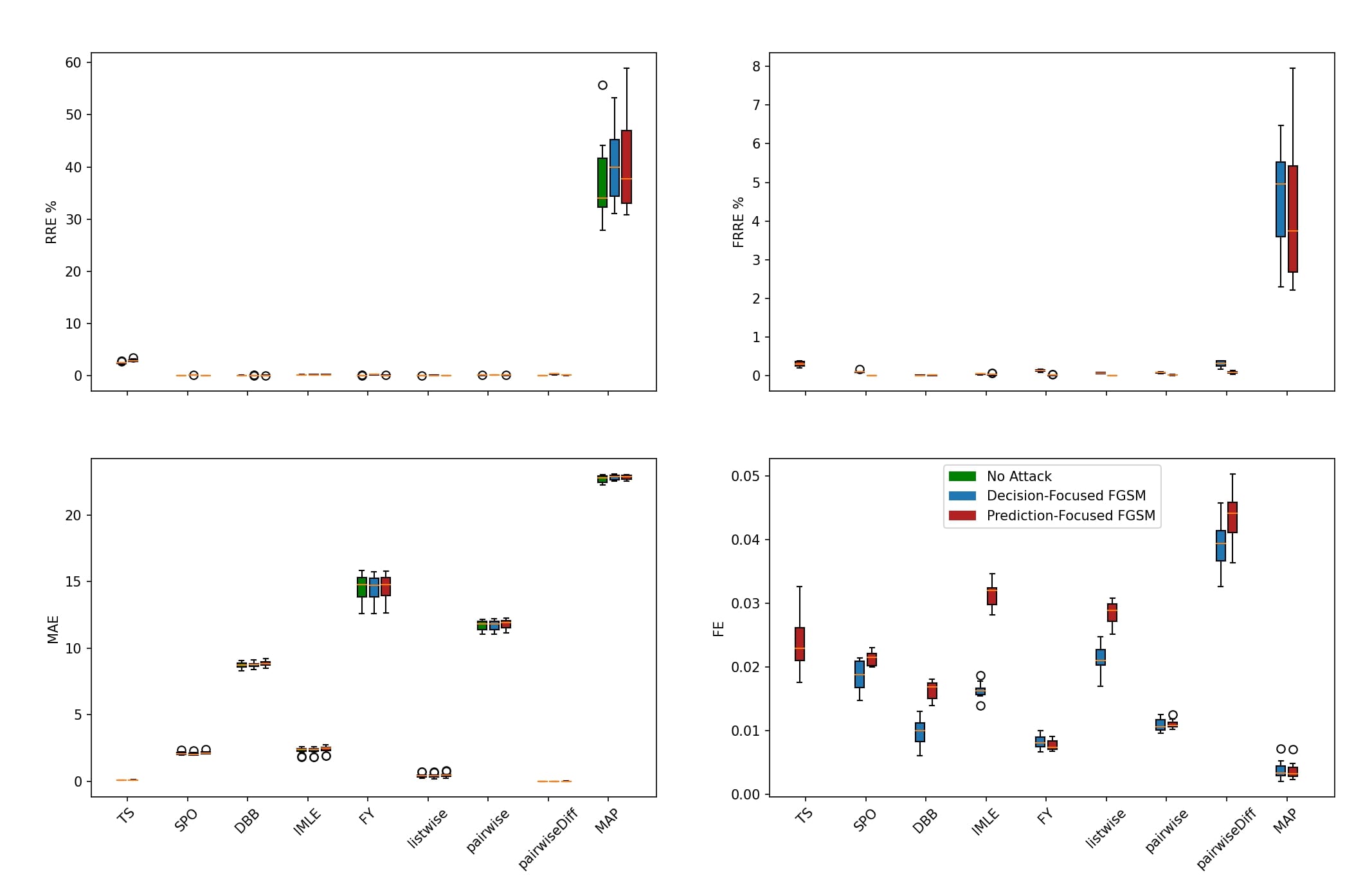} 
\caption{WarCraft img size $12 \times 12$, Perturbation Magnitude 0.01}
\label{W12_0.01}
\end{figure}

\begin{figure}[htbp]
\centering
\includegraphics[width = 8cm, height = 5cm]{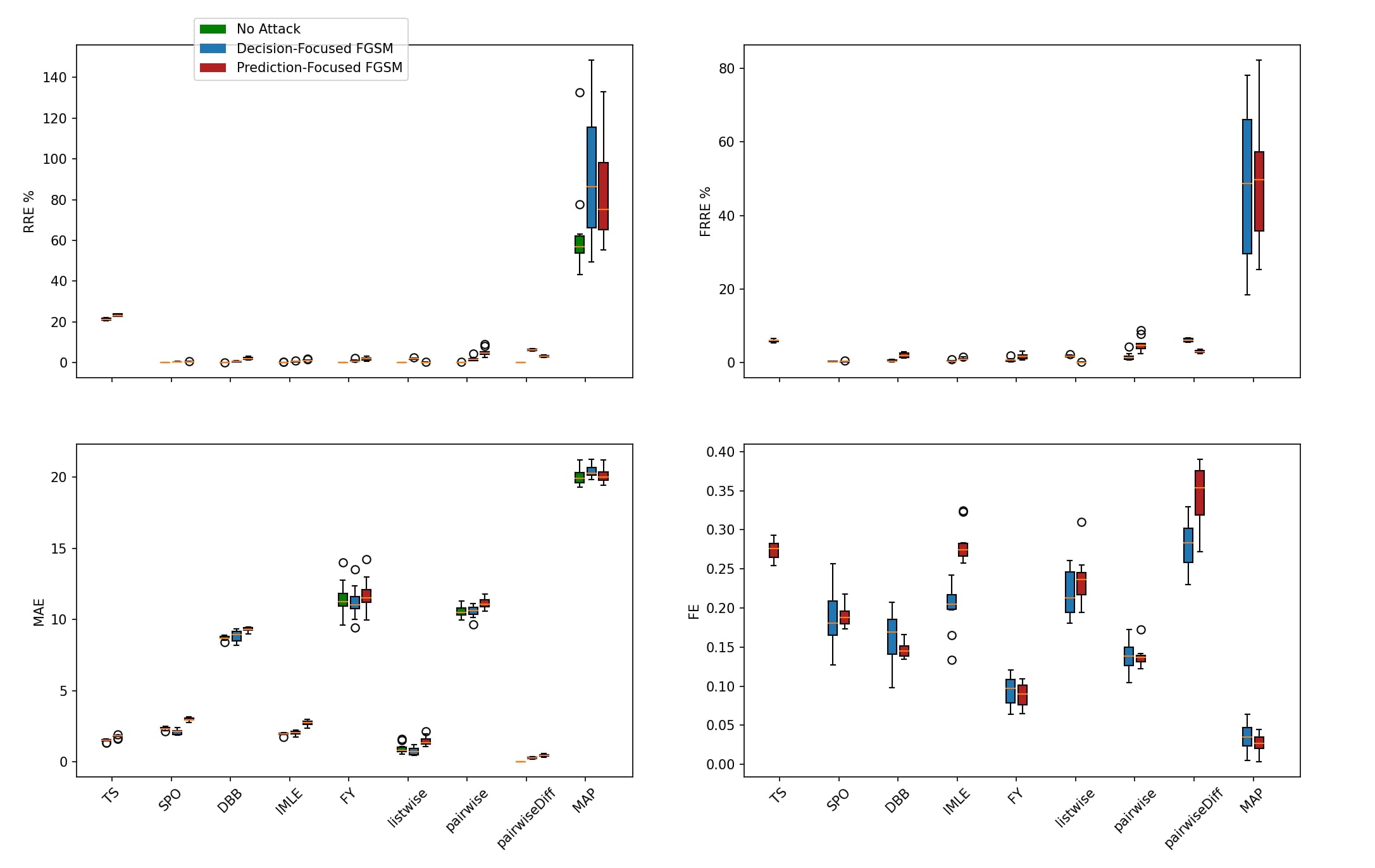} 
\caption{WarCraft img size $24 \times 24$, Perturbation Magnitude 0.15}
\label{W24_0.15}
\end{figure}

\begin{figure}[htbp]
\centering
\includegraphics[width = 8cm, height = 5cm]{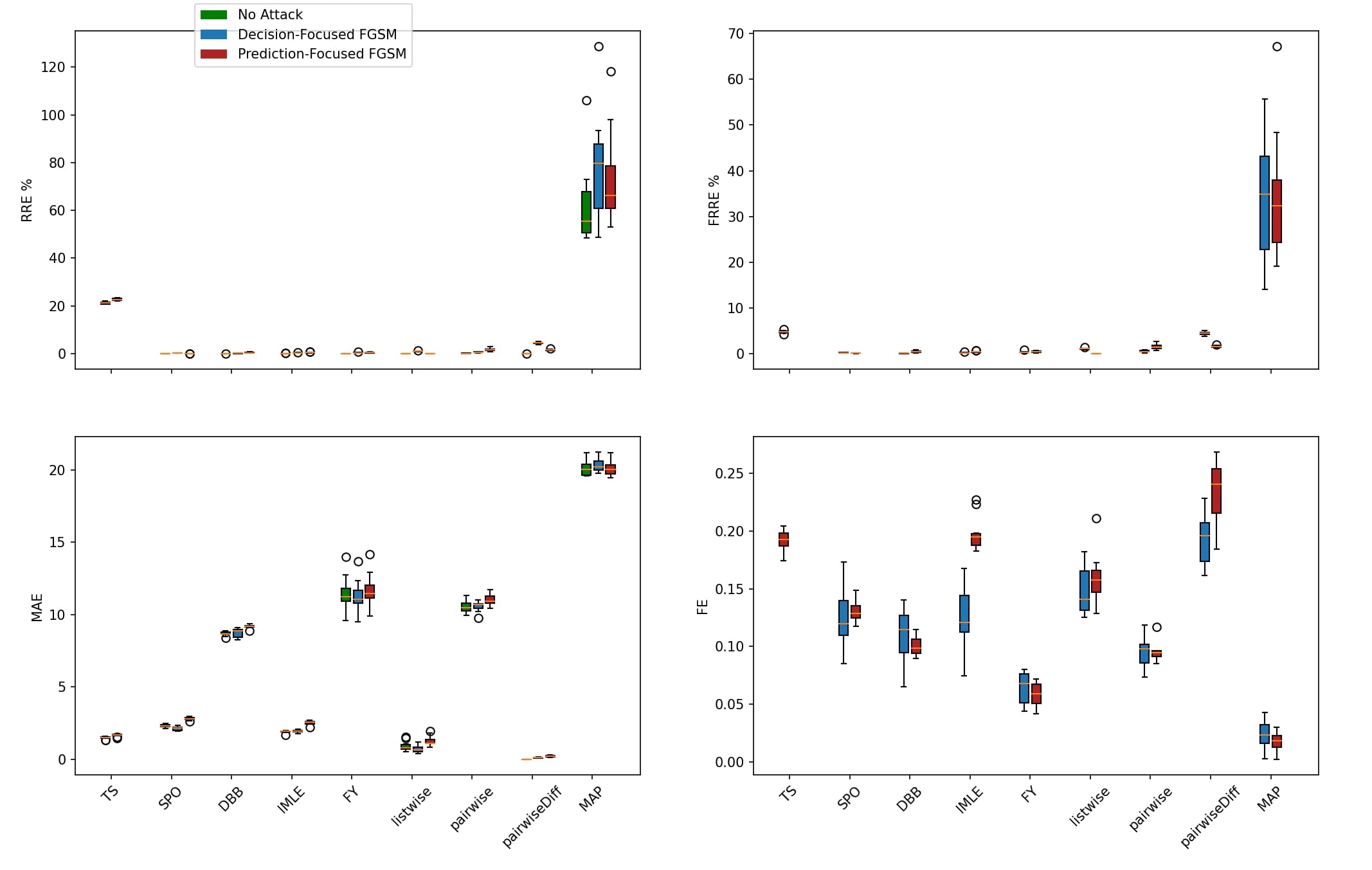} 
\caption{WarCraft img size $24 \times 24$, Perturbation Magnitude 0.1}
\label{W24_0.1}
\end{figure}

\begin{figure}[htbp]
\centering
\includegraphics[width = 8cm, height = 5cm]{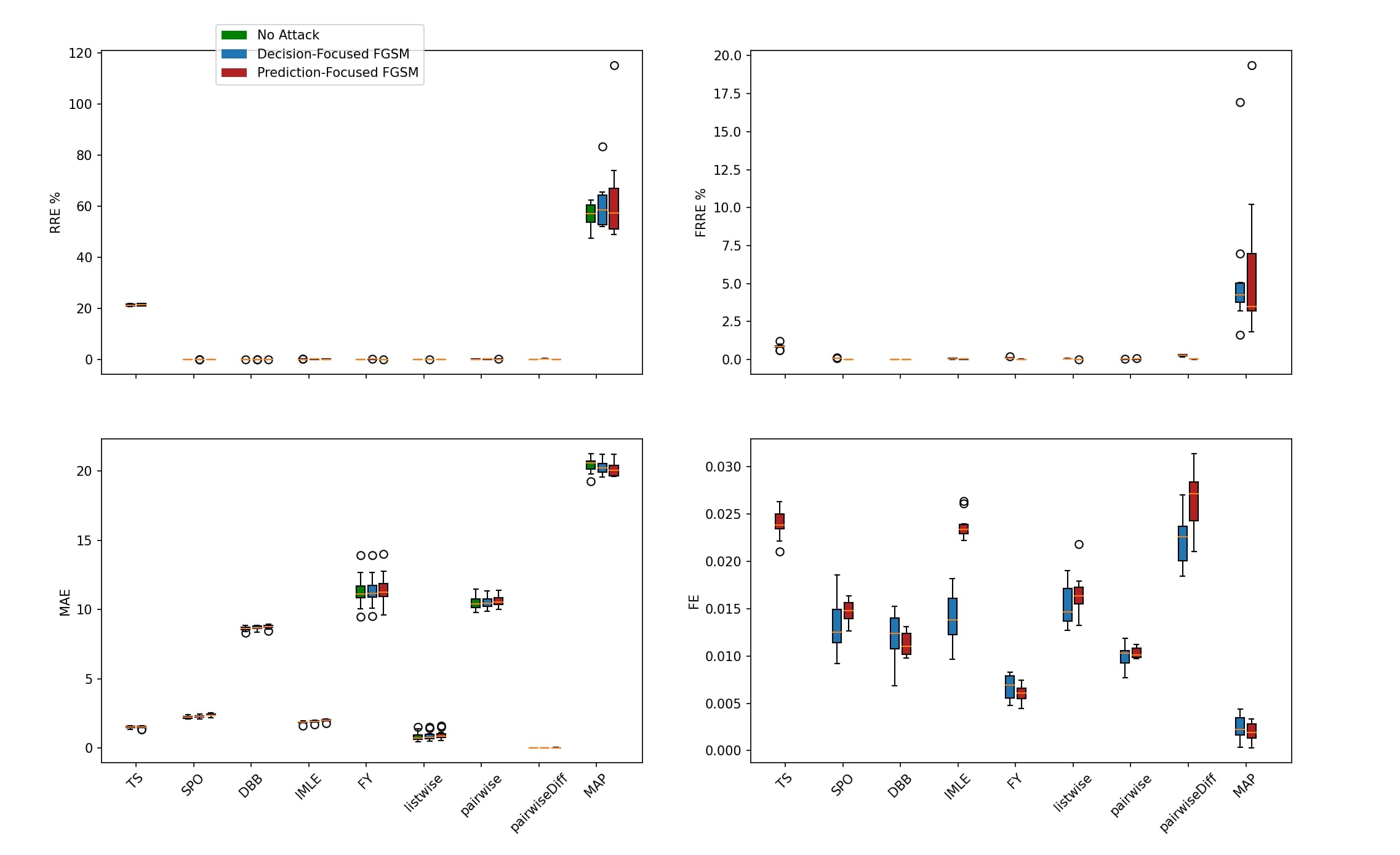} 
\caption{WarCraft img size $24 \times 24$, Perturbation Magnitude 0.01}
\label{W24_0.01}
\end{figure}

\begin{figure}[htbp]
\centering
\includegraphics[width = 8cm, height = 5cm]{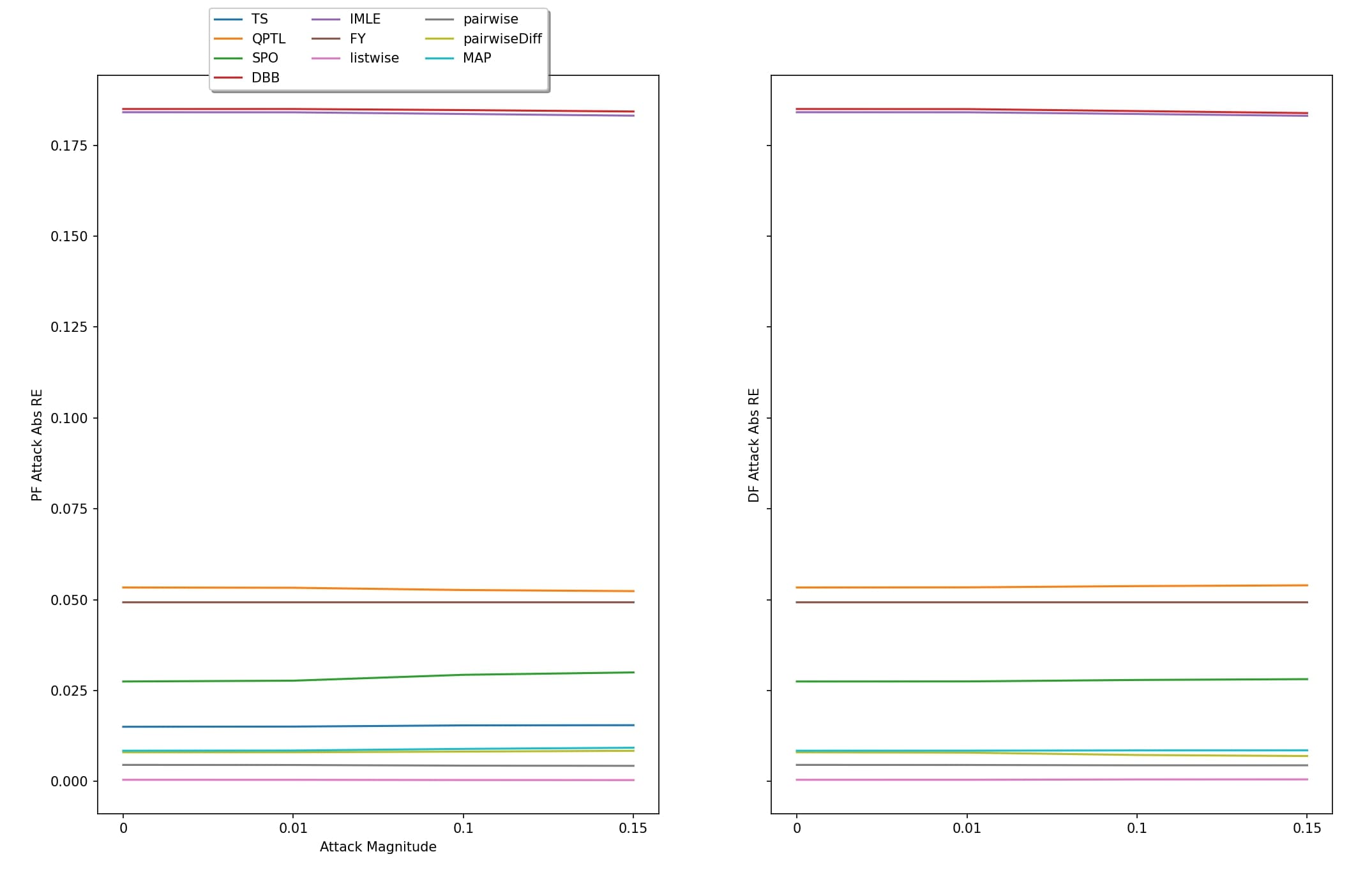} 
\caption{Portfolio deg 16, Absolute RE}
\label{AP16_AbsRE}
\end{figure}

\begin{figure}[htbp]
\centering
\includegraphics[width = 8cm, height = 5cm]{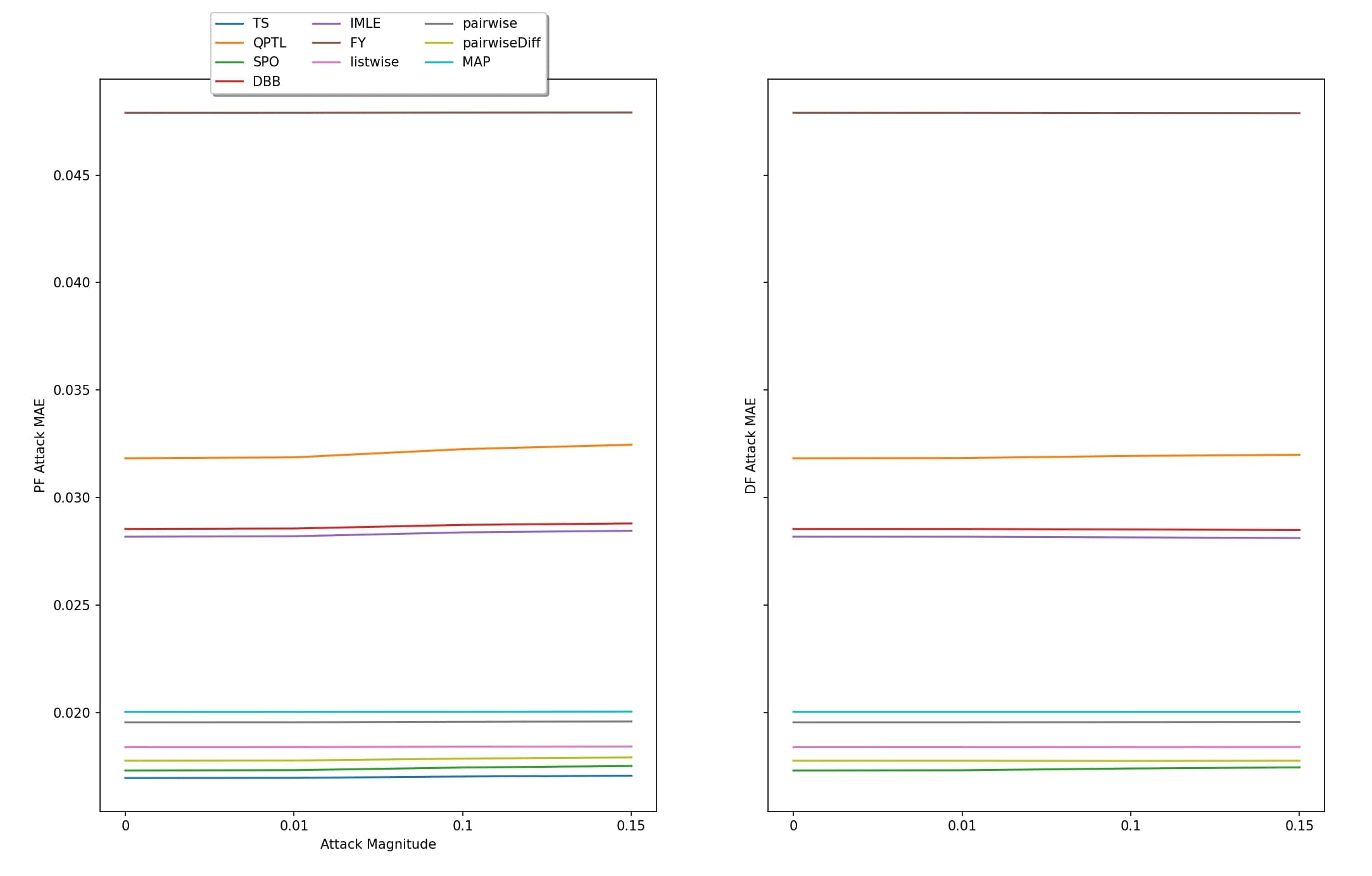} 
\caption{Portfolio deg 16, MAE}
\label{AP16_MAE}
\end{figure}

\begin{figure}[htbp]
\centering
\includegraphics[width = 8cm, height = 5cm]{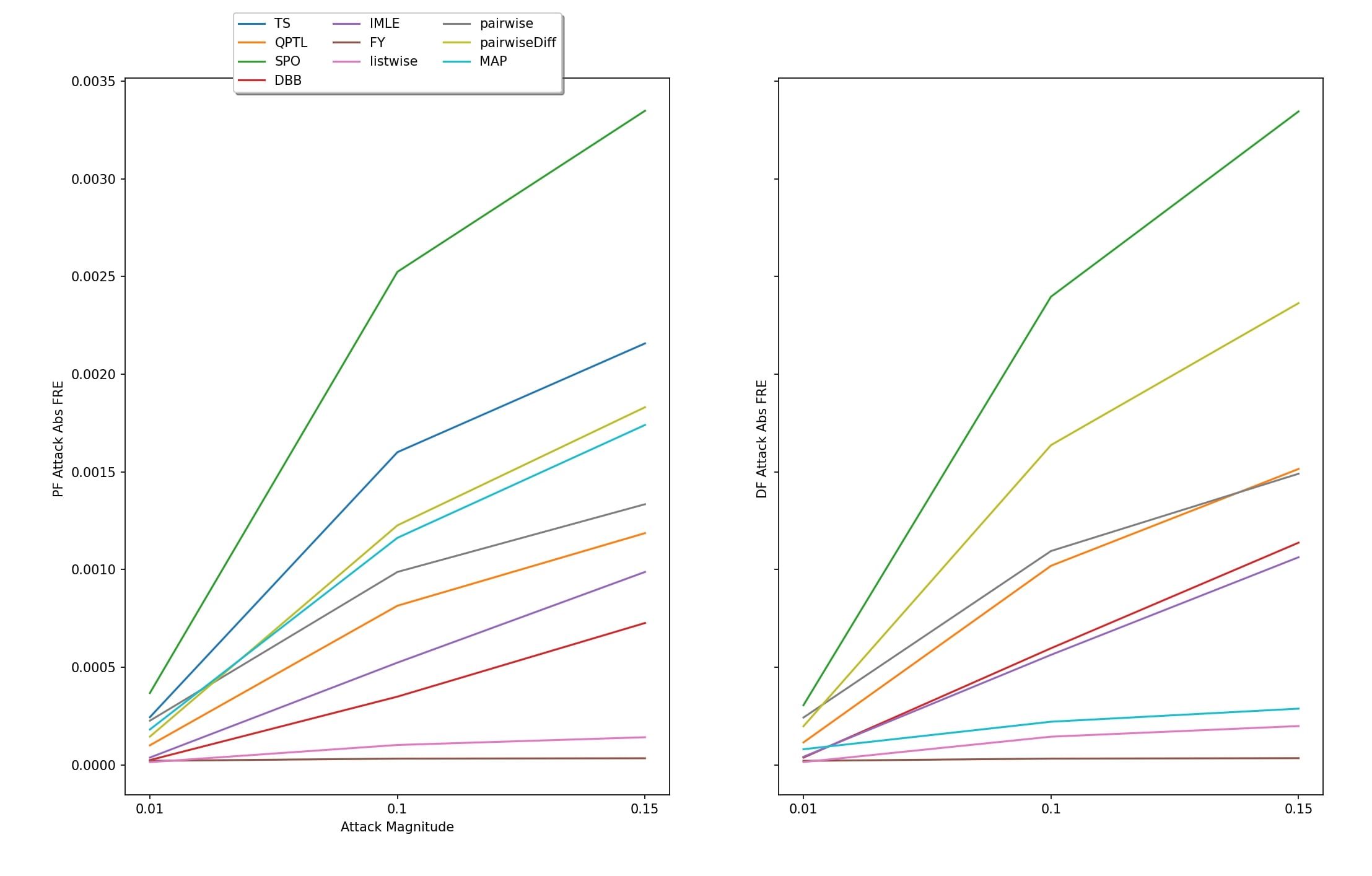} 
\caption{Portfolio deg 16, Absolute FRE}
\label{AP16_AbsFRE}
\end{figure}

\begin{figure}[htbp]
\centering
\includegraphics[width = 8cm, height = 5cm]{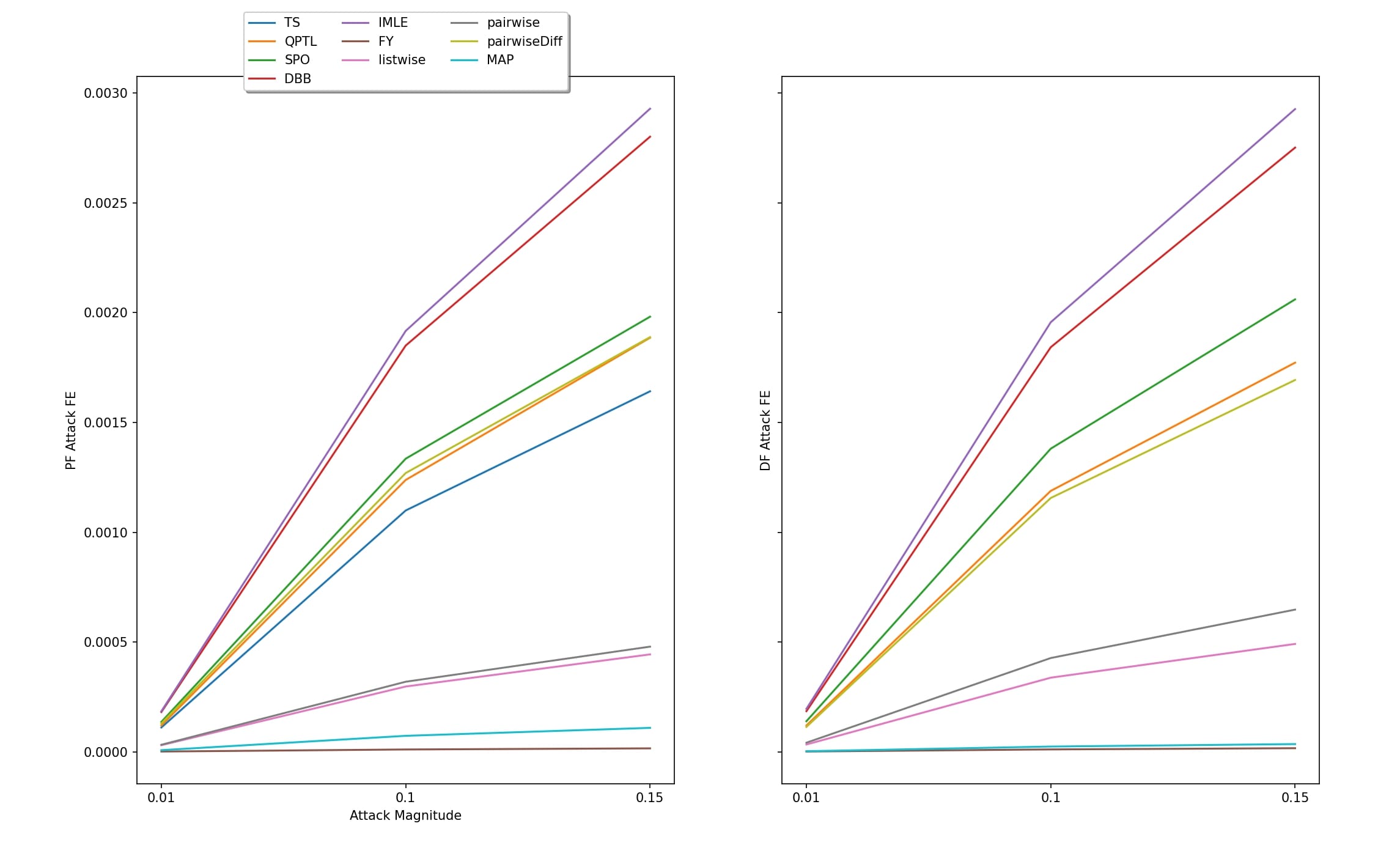} 
\caption{Portfolio deg 16, FE}
\label{AP16_FE}
\end{figure}

\begin{figure}[htbp]
\centering
\includegraphics[width = 8cm, height = 5cm]{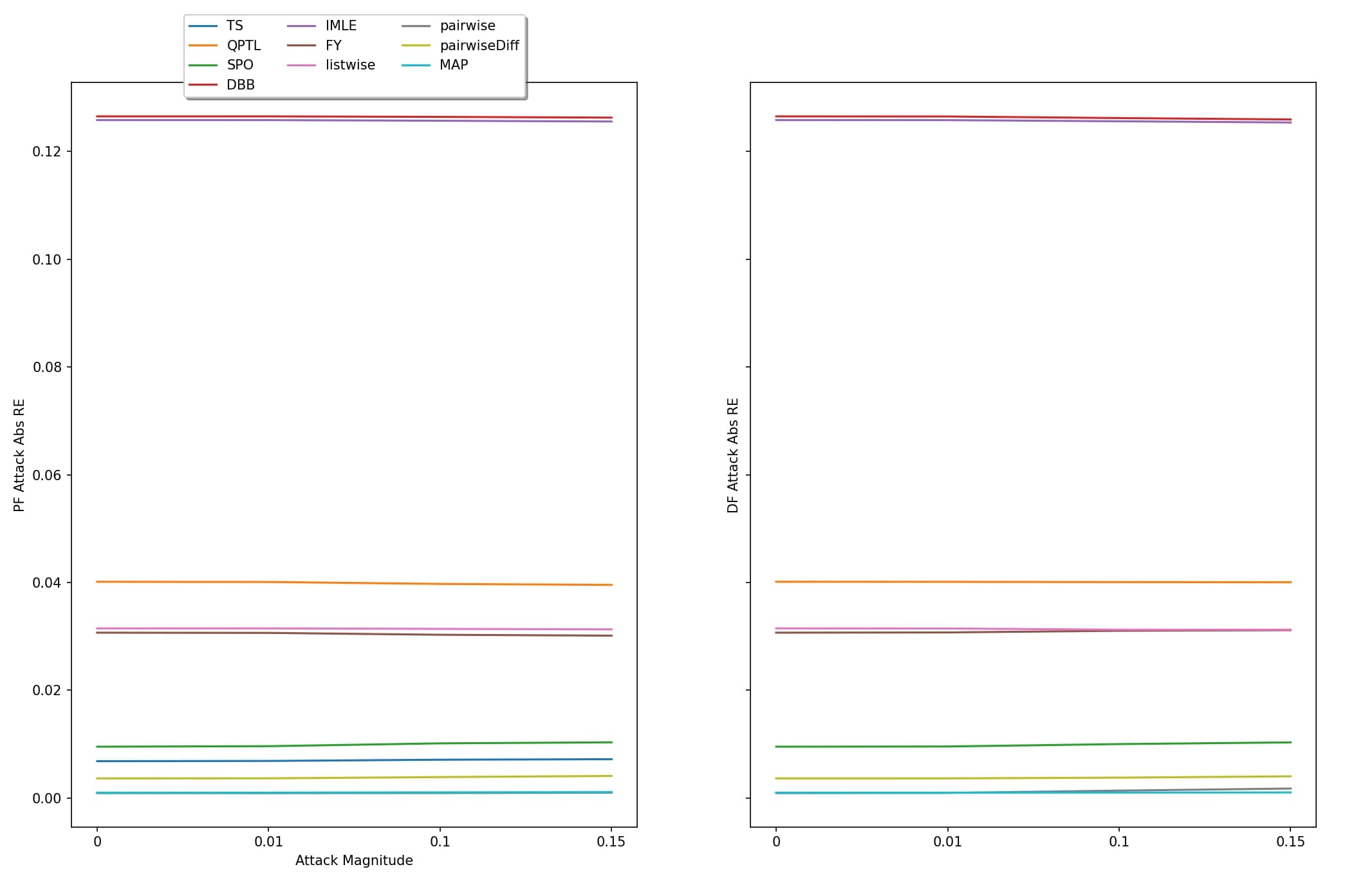} 
\caption{Portfolio deg 1, Absolute RE}
\label{AP1_AbsRE}
\end{figure}

\begin{figure}[htbp]
\centering
\includegraphics[width = 8cm, height = 5cm]{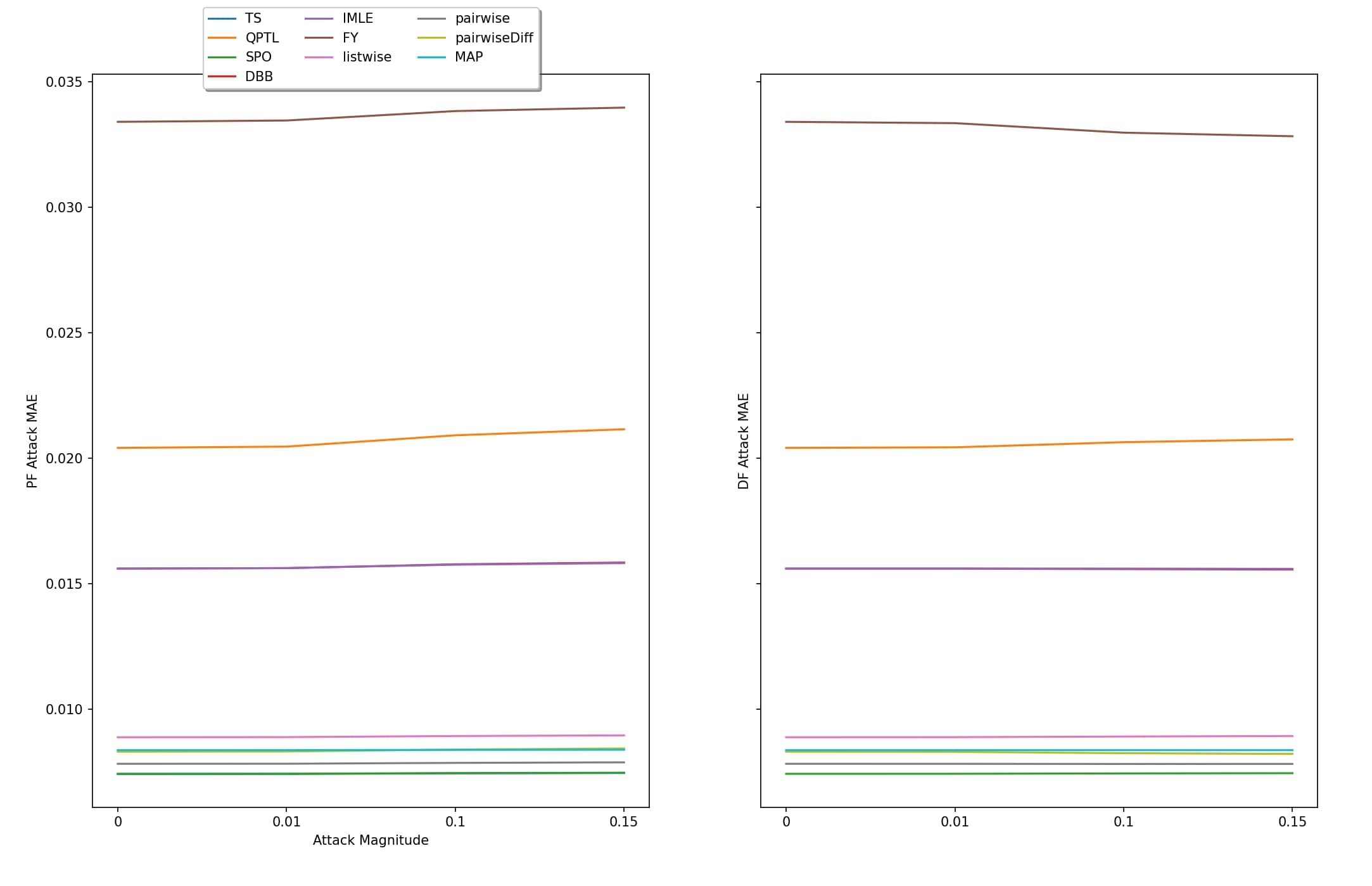} 
\caption{Portfolio deg 1, MAE}
\label{AP1_MAE}
\end{figure}

\begin{figure}[htbp]
\centering
\includegraphics[width = 8cm, height = 5cm]{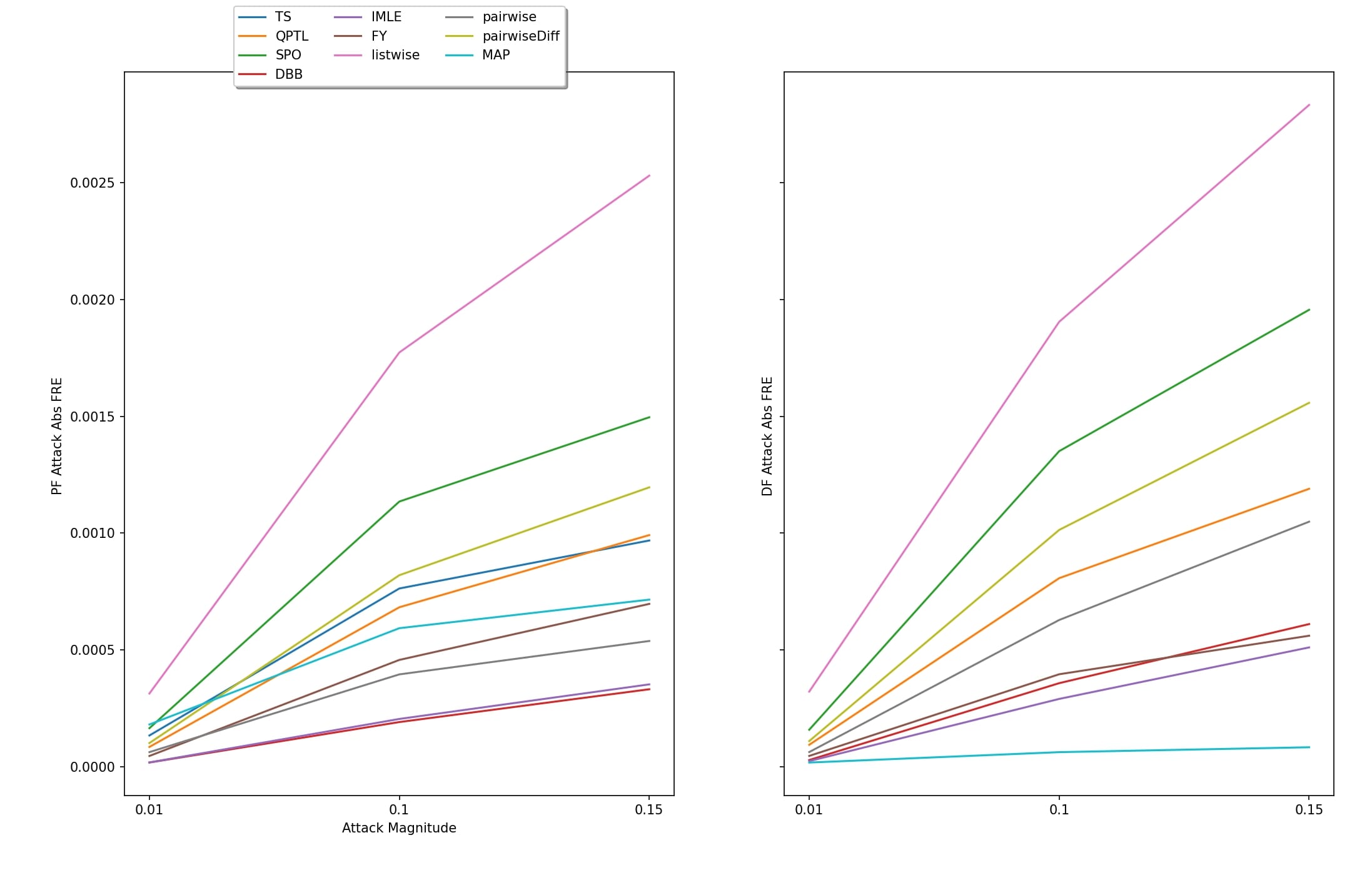} 
\caption{Portfolio deg 1, Absolute FRE}
\label{AP1_AbsFRE}
\end{figure}

\begin{figure}[htbp]
\centering
\includegraphics[width = 8cm, height = 5cm]{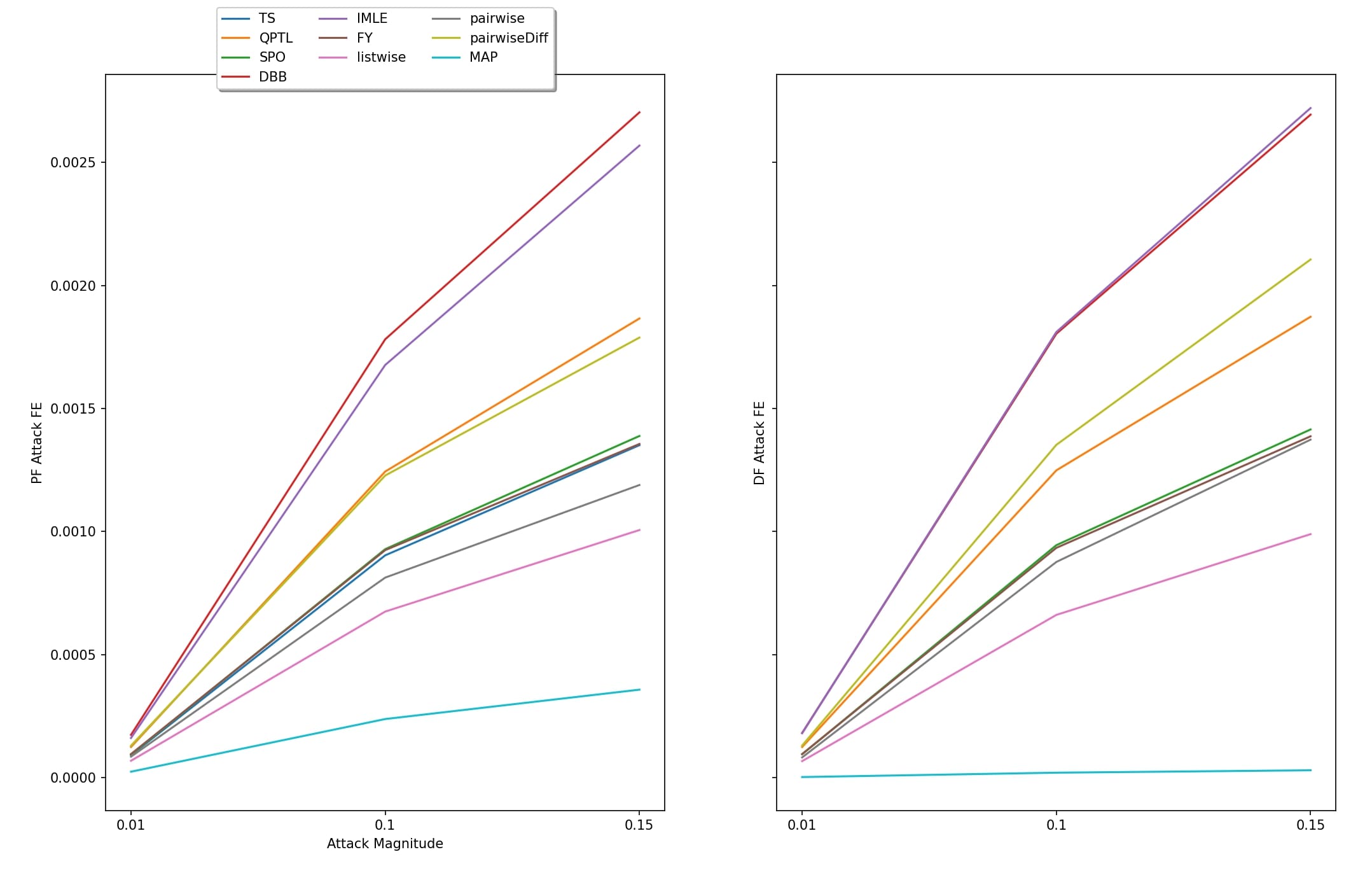} 
\caption{Portfolio deg 1, FE}
\label{AP1_FE}
\end{figure}


\begin{figure}[htbp]
\centering
\includegraphics[width = 8cm, height = 5cm]{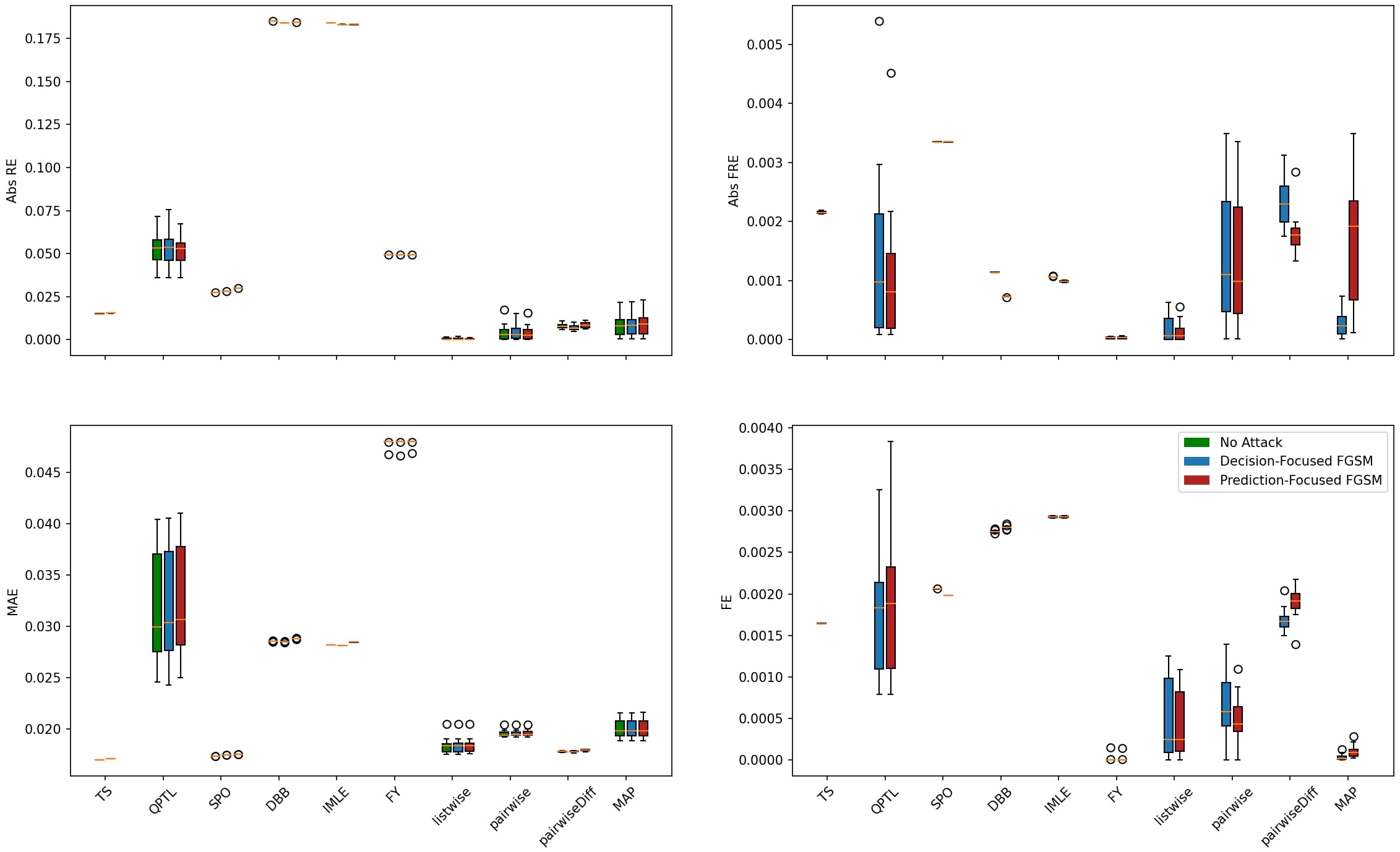} 
\caption{Portfolio Degree 16, Perturbation Magnitude 0.15}
\label{P16_0.15}
\end{figure}

\begin{figure}[htbp]
\centering
\includegraphics[width = 8cm, height = 5cm]{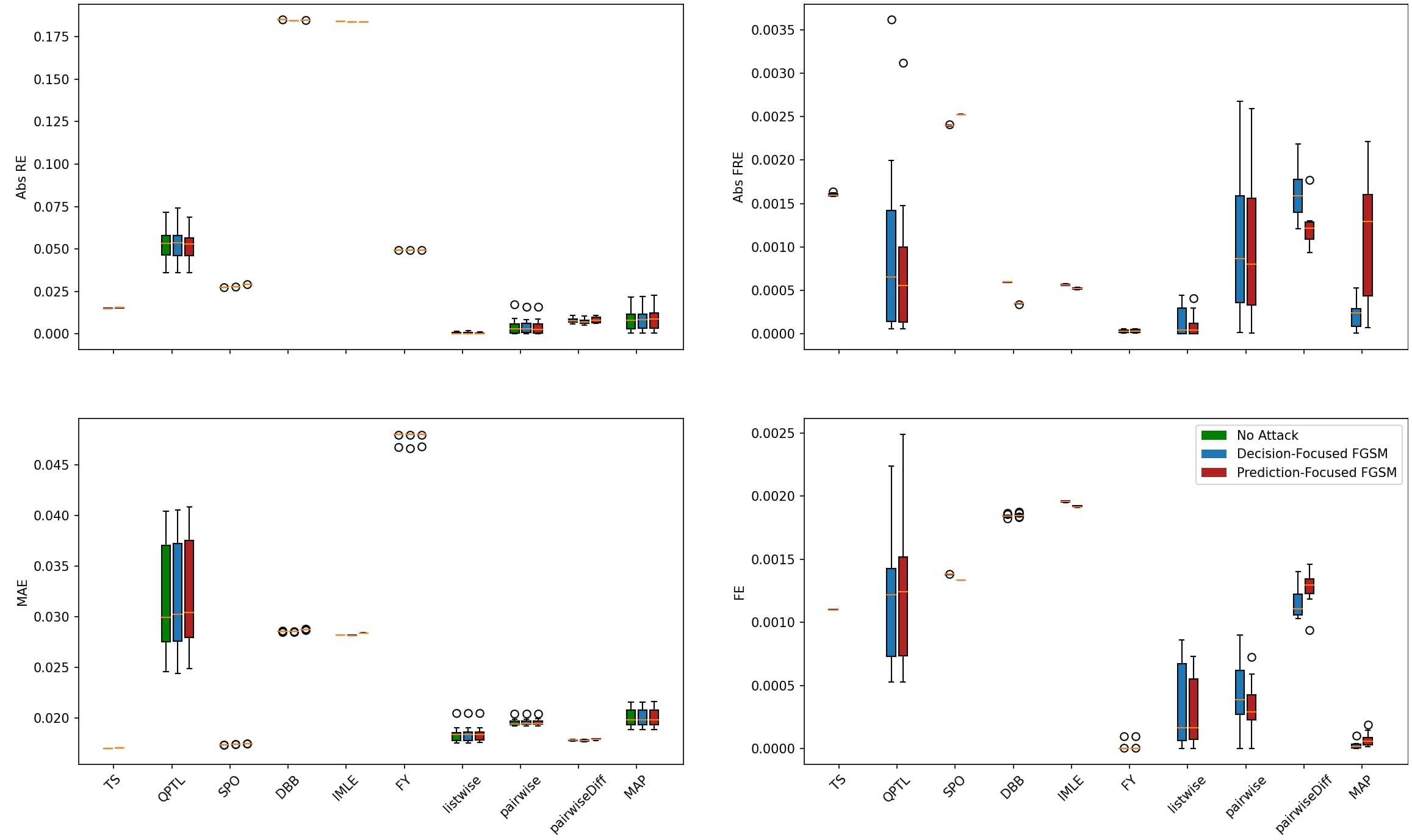} 
\caption{Portfolio Degree 16, Perturbation Magnitude 0.1}
\label{P16_0.1}
\end{figure}

\begin{figure}[htbp]
\centering
\includegraphics[width = 8cm, height = 5cm]{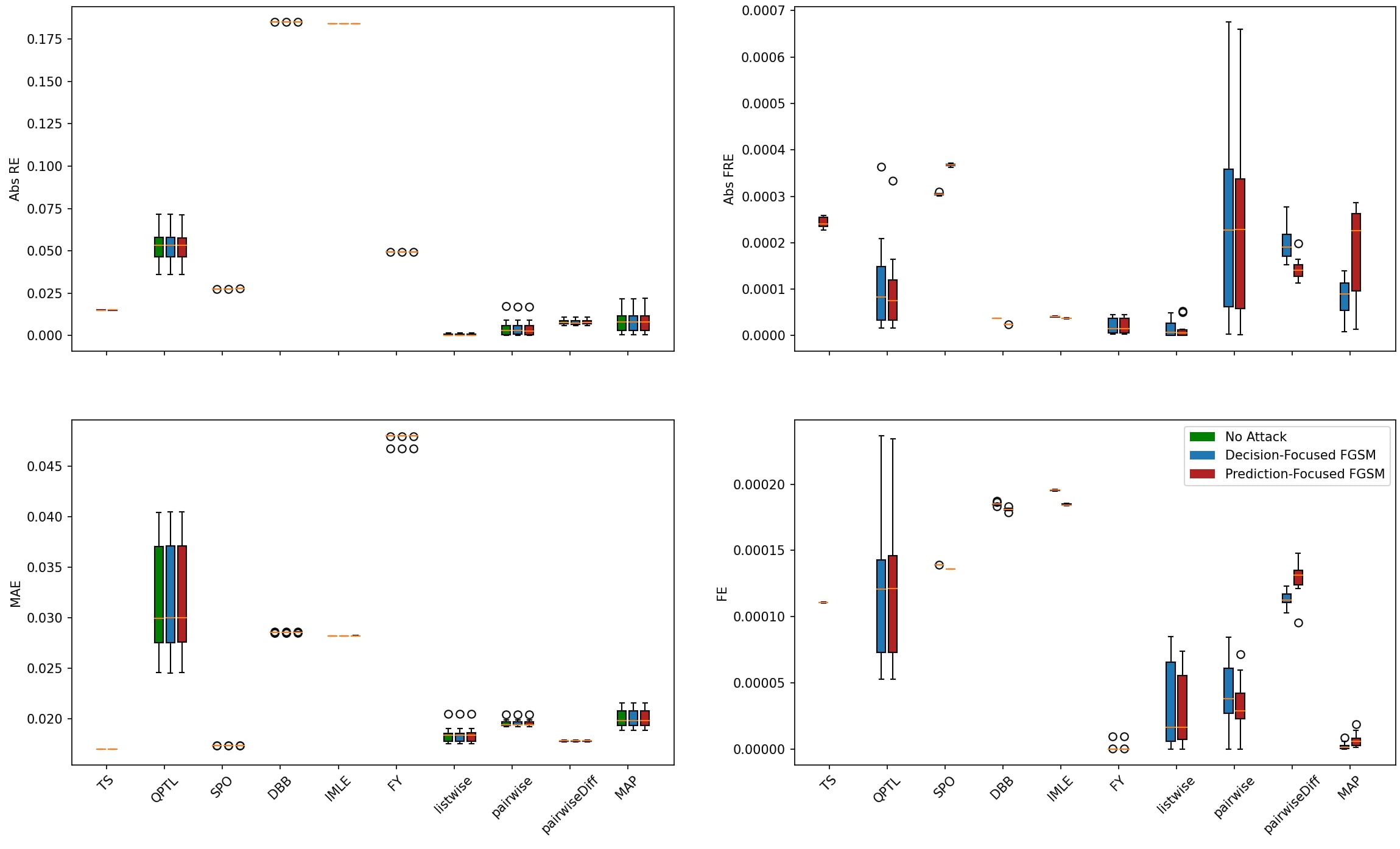} 
\caption{Portfolio Degree 16, Perturbation Magnitude 0.01}
\label{P16_0.01}
\end{figure}

\begin{figure}[htbp]
\centering
\includegraphics[width = 8cm, height = 5cm]{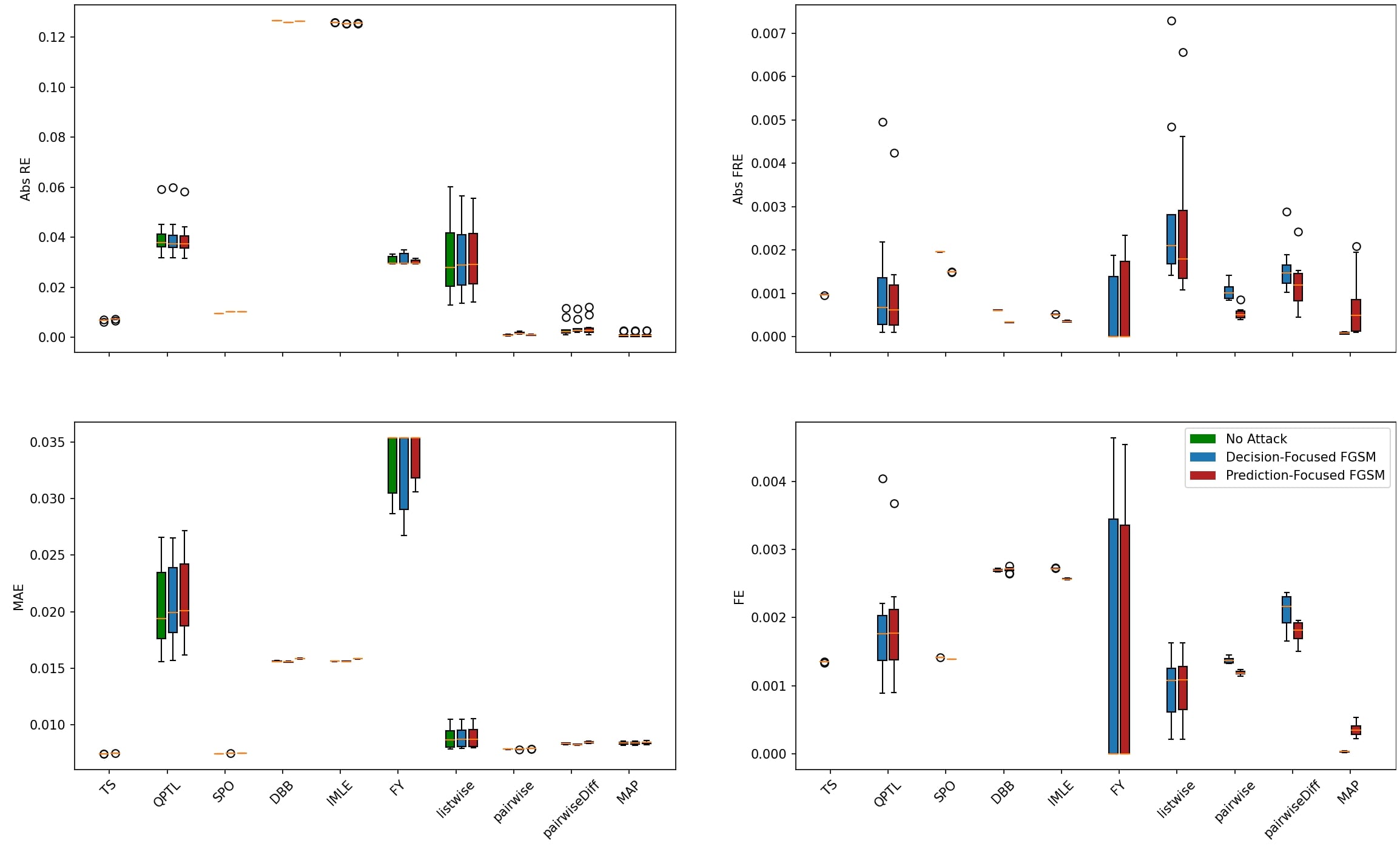} 
\caption{Portfolio Degree 1, Perturbation Magnitude 0.15}
\label{P1_0.15}
\end{figure}

\begin{figure}[htbp]
\centering
\includegraphics[width = 8cm, height = 5cm]{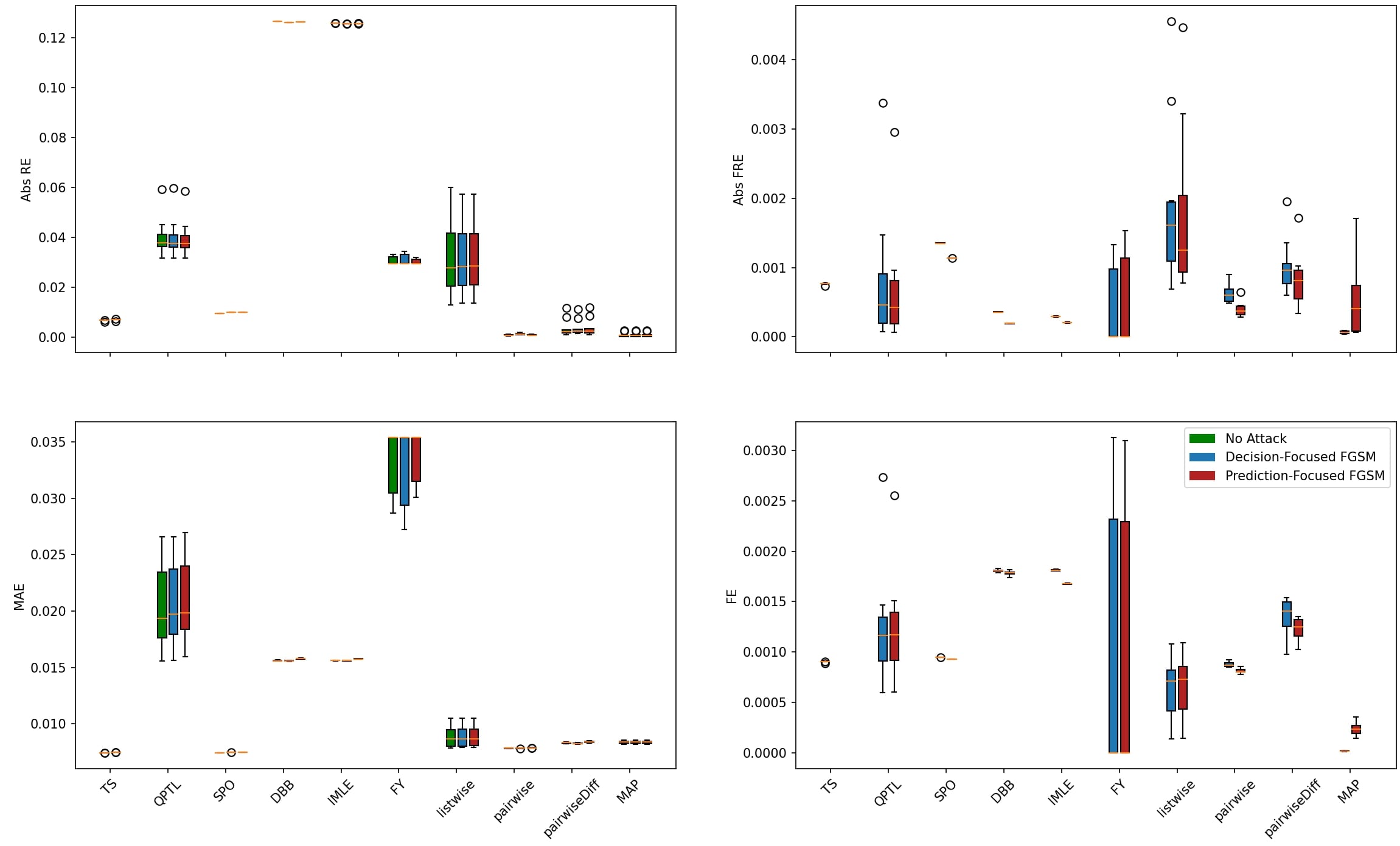} 
\caption{Portfolio Degree 1, Perturbation Magnitude 0.1}
\label{P1_0.1}
\end{figure}

\begin{figure}[htbp]
\centering
\includegraphics[width = 8cm, height = 5cm]{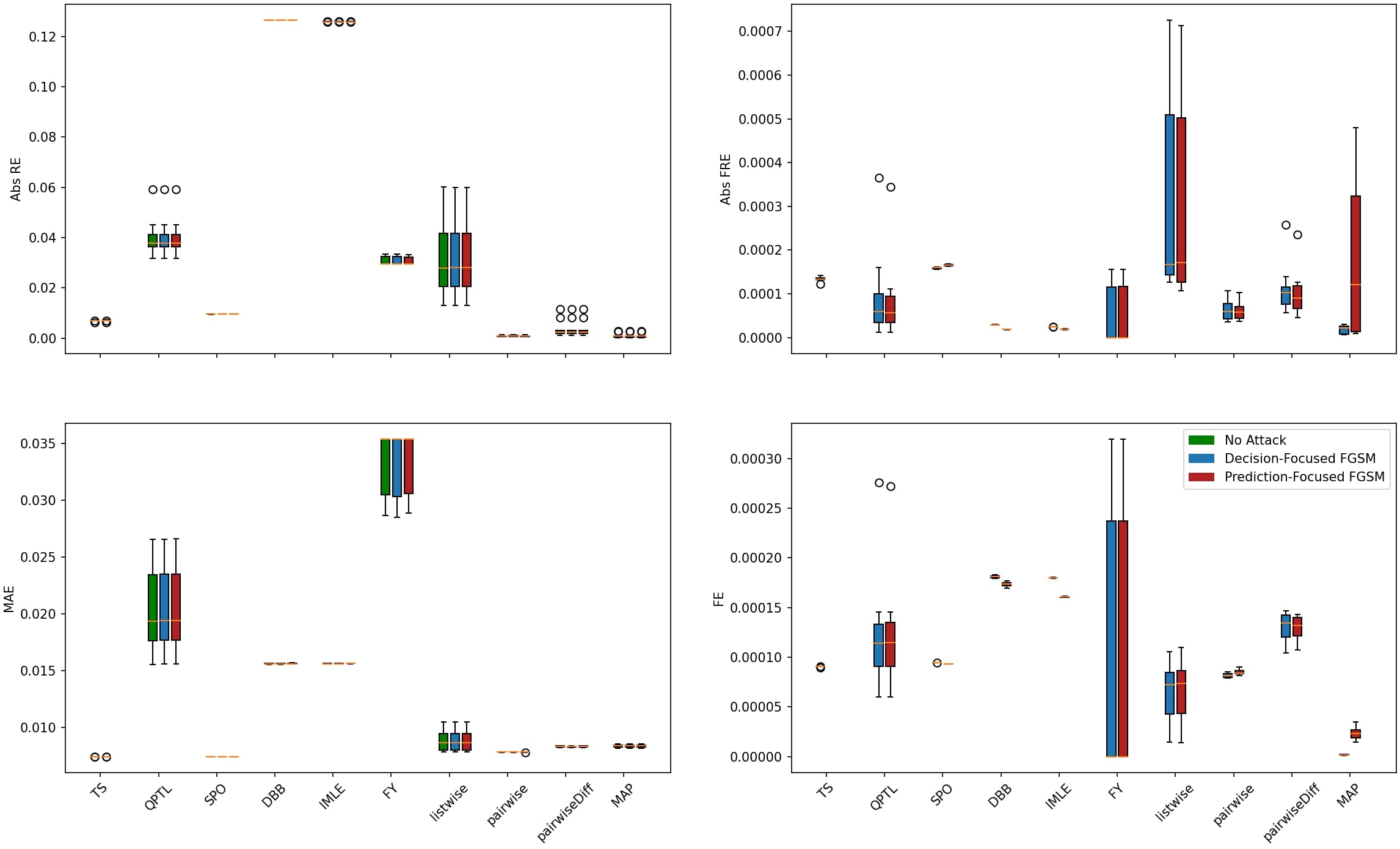} 
\caption{Portfolio Degree 1, Perturbation Magnitude 0.01}
\label{P1_0.01}
\end{figure}

\section*{Acknowledgments}
Special thanks to Dr. Ferdinando Fioretto for the direction and support of this work, and to James Kotary for the interesting conversations and insights.

\section*{Disclosure Statement} The authors declare that there are no relevant financial or non-financial competing interests associated with this manuscript.

\bibliography{aaai24}

\end{document}